%% file: main.tex
\documentclass[10pt]{article}

\input{sections/preamble}

\title{A Decision-driven Methodology for Designing Uncertainty-aware AI Self-Assessment}

\author[1]{Gregory Canal}
\author[1]{Vladimir Leung}
\author[1]{Philip Sage}
\author[2]{Eric Heim}
\author[1]{I-Jeng Wang}
\affil[1]{The Johns Hopkins University Applied Physics Laboratory\protect\\
	\texttt{\{Greg.Canal,Vladimir.Leung,Philip.Sage,I-Jeng.Wang\}@jhuapl.edu}}
\affil[2]{Software Engineering Institute\protect\\
	Carnegie Mellon University\protect\\
	\texttt{etheim@sei.cmu.edu}}

\date{}

\begin{document}
	

\maketitle

\begin{abstract}
\input{sections/abstract}
\end{abstract}

\input{sections/introduction}
\input{sections/attributes}
\input{sections/overview}
\input{sections/notional_examples}

\input{sections/discussion}

\section*{Acknowledgement}
This work was supported by the Office of the Under Secretary of Defense, Research and Engineering (OUSD (R\&E)). 

\printbibliography



\end{document}

%% file: sections/preamble.tex
\usepackage[margin=1in]{geometry}
\usepackage{authblk}
\usepackage{booktabs}
\usepackage{microtype}
\usepackage{hyperref}
\usepackage[style=ieee]{biblatex}

\usepackage{graphicx}
\usepackage{wrapfig}
\usepackage{caption, subcaption}
\usepackage{placeins}

\usepackage{amsmath, amssymb, amsthm, amsfonts, bm, bbm}
\usepackage{dsfont} 
\usepackage{mathtools}
\usepackage{algorithm}
\usepackage{algorithmic}
\usepackage{nicefrac}
\usepackage{array}

\usepackage[normalem]{ulem} 
\usepackage{xspace}
\usepackage{multirow}
\usepackage{makecell}
\usepackage{hhline}

\usepackage{xcolor} 

\usepackage{cleveref}
\usepackage[T1]{fontenc}
\usepackage{lmodern}

\usepackage{enumitem,amssymb}
\newlist{todolist}{itemize}{2}
\setlist[todolist]{label=$\square$}
\usepackage{pifont}
\DeclareMathOperator*{\argmax}{arg\,max}
\DeclareMathOperator*{\argmin}{arg\,min}

\DeclarePairedDelimiter{\abs}{\lvert}{\rvert}

\DeclarePairedDelimiterX{\ip}[2]{\langle}{\rangle}{#1, #2}
\DeclarePairedDelimiterX{\KLx}[2]{(}{)}{%
	#1\:\delimsize\|\:#2%
}

\DeclareMathOperator{\E}{\mathbb{E}}

\newcommand{\R}{\mathbb{R}}

\newcommand{\bb}{\bm{b}}

\newcommand{\bW}{\bm{W}}

\newcommand{\cA}{\mathcal{A}}

\newcommand{\cD}{\mathcal{D}}

\newcommand{\cN}{\mathcal{N}}

\newcommand{\cS}{\mathcal{S}}

\newcommand{\cW}{\mathcal{W}}
\newcommand{\cX}{\mathcal{X}}
\newcommand{\cY}{\mathcal{Y}}
\newcommand{\cZ}{\mathcal{Z}}

\newcommand{\eps}{\varepsilon}

\newcommand{\ind}{\mathds{1}}

\theoremstyle{remark}

\theoremstyle{definition}

\graphicspath{{.}{./figures}} 

%% file: sections/abstract.tex
Artificial intelligence (AI) has revolutionized decision-making processes and systems throughout society and, in particular, has emerged as a significant technology in high-impact scenarios of national interest. Yet, despite AI's impressive predictive capabilities in controlled settings, it still suffers from a range of practical setbacks preventing its widespread use in various critical scenarios. In particular, it is generally unclear if a given AI system's predictions can be \emph{trusted} by decision-makers in downstream applications. To address the need for more transparent, robust, and trustworthy AI systems, a suite of tools has been developed to quantify the uncertainty of AI predictions and, more generally, enable AI to ``self-assess'' the reliability of its predictions. In this manuscript, we categorize methods for AI self-assessment along several key dimensions and provide guidelines for selecting and designing the appropriate method for a practitioner's needs. In particular, we focus on uncertainty estimation techniques that consider the impact of self-assessment on the choices made by downstream decision-makers and on the resulting costs and benefits of decision outcomes. To demonstrate the utility of our methodology for self-assessment design, we illustrate its use for two realistic national-interest scenarios. This manuscript is a practical guide for machine learning engineers and AI system users to select the ideal self-assessment techniques for each problem.\footnote{Approved for open publication by the U.S. Department of Defense Office of Publication and Security Review on July 30, 2024.}

%% file: sections/introduction.tex
\section{Introduction}
\label{sec:intro}

In the past decade, the world has witnessed an explosion in the capabilities of artificial intelligence (AI) systems, and their use has proliferated throughout most corners of society. AI systems have been deployed in applications ranging from commercial and industrial uses, such as recommendation systems for social networks, advertising, and e-commerce, to platforms of national interest, including defense, healthcare, climate, and scientific sectors. While the power of generative and predictive AI systems to imitate, augment, and enhance human capabilities has become abundantly clear, arguably the most significant roadblock to fully deploying AI systems at scale in critical problem scenarios has been a lack of certainty about the reliability, robustness, and trustworthiness of their predictions and model outputs.

In general, AI systems can have their performance limited by a range of practical considerations, such as reproducing systematic bias present in their training data, failing to generalize to domains outside those encountered during training, overfitting to spurious correlations present in data, or lacking the ability to abstain from making unfounded predictions. In certain commercial applications such as online streaming, such predictive flaws are only moderately problematic since, in the worst case, a user may be provided with a suboptimal user experience (e.g., being recommended movies they would not typically enjoy) rather than suffering a catastrophic loss. However, in many problems of national interest, such shortcomings cannot be tolerated due to the high-impact nature of such settings, the large sway that AI predictions might have on downstream decisions, and the potential for poor decisions to cause catastrophic failures or significant opportunity costs.

To more fundamentally address shortcomings in the trustworthiness of AI systems, researchers and practitioners alike have developed various tools and techniques to increase AI transparency, reliability, and robustness. One general class of such approaches relies on a separation between two types of AI outputs: the specific \emph{predictions} made by an AI system for a particular input and an associated \emph{self-assessment} (SeA) measuring the AI's confidence in its predictions. By providing such a self-assessment, the AI makes a downstream decision-maker more fully aware of any uncertainty or sources of ambiguity in the AI predictions. In general, there are innumerable forms of self-assessment that an AI system might provide, and the set of techniques and literature satisfying the definition of self-assessment is vast, potentially including broad fields such as explainable and interpretable AI \cite{dwivedi2023explainable, minh2022explainable}.

One prominent category of self-assessment approaches are those methods that \emph{quantify the degree of uncertainty} an AI has in its predictions (which we denote as \emph{uncertainty-aware} self-assessment), allowing downstream decision-makers to weigh AI predictions by their associated confidence levels. In general, the confidence outputs supplied by self-assessment can drastically affect downstream decisions, depending on how the decision-maker takes these weights into account, along with the potential costs and risks associated with each candidate decision. For instance, a human decision-maker may only move forward with high-risk actions if the AI's associated confidence level is above a particular threshold. Similarly, self-assessment confidence levels can  have a significant impact on algorithmic (non-human) decision-makers since optimal decision policies typically weigh each potential decision cost by the AI's estimated uncertainty and select the action with the smallest weighted cost. Through this lens, AI self-assessment techniques are a crucial component in the overall decision-making pipeline and should be selected and designed in a \emph{decision-driven} manner.

As an illustrative example of how self-assessment uncertainty outputs can impact downstream decision costs, consider the following scenario: suppose a decision-maker is playing a game where she is given \$20 and then presented with a closed box that either contains \$100 or is empty. The decision-maker is aware of both possibilities but does not know which amount the box contains until it is opened. She is given the choice to keep her \$20 and end the game (leaving the box closed) or pay back the \$20 for the chance to open the box and keep whatever value is inside. If she opens the box and it contains \$100, she will walk away with a net profit of \$80, but if the box is empty, she will have lost her \$20 and walk away empty-handed (net profit of \$0). Although looking at the box gives her no information, she is aided by an AI that predicts whether the box contains the \$100 and also provides a self-assessment confidence level $c$ between 0-100\%.

If the decision-maker assumes that this confidence level is well-calibrated, she can calculate her expected profit from opening the box to be $\$80 \times c$. On the other hand, if she does not open the box and walks away, her net profit from playing the game is always $\$20$. Based on this calculation, opening the box is only worth the risk if $\$80 \times c > \$20$, which occurs when the AI's confidence $c$ is at least $25\%$. If the decision-maker follows this logic, she will decide to pay \$20 to open the box if the AI thinks it contains \$100 with confidence at least 25\%, and will walk away with her \$20 otherwise. Hence, the AI's self-assessed confidence level can have a tremendous impact on the decision process since the difference of a few confidence percentage points around 25\% might result in a different downstream action. Conversely, in this context, the AI's confidence level does not need to be well-calibrated outside of this small percentage range since all that matters for decision-making is if the confidence is above or below 25\% rather than its exact value.

\textbf{Our Contribution:} this manuscript presents a decision-driven methodology for selecting and designing uncertainty-aware AI self-assessment techniques. Although previous surveys and general frameworks related to self-assessment have been developed \cite{yang2024generalized, abdar2021review, gawlikowski2023survey, ghosh2021uncertainty, kirchenbauer2022what, chung2021uncertainty}, to our knowledge no previous surveys emphasize the specific impact of self-assessment on downstream decisions and their associated costs, as we do here. Unlike some surveys, the intention of this manuscript is to provide \emph{practical guidance} for machine learning (ML) engineers and system users to select between and evaluate various self-assessment methods. In \Cref{sec:attr}, we provide an overview of the key attributes we use to categorize various self-assessment methods along with a set of guidelines for how a user might select appropriate self-assessment techniques for their problem, taking into consideration the nuances and constraints of their AI pipeline and downstream decision-making process.

In \Cref{sec:overview}, we detail a range of self-assessment techniques that a practitioner might utilize and specifically categorize these based on their level of decision awareness. We note that these candidate methods should be viewed as a representative set of techniques in the literature rather than an exhaustive list, as the number of available uncertainty quantification techniques continues to grow rapidly. Additionally, these presented methods should not be construed as a fixed list of methods for a practitioner to select a single method from. Instead, practitioners should utilize the framework presented in \Cref{sec:attr} to formulate the ideal self-assessment for their application, and take inspiration from the methods presented in \Cref{sec:overview}, possibly combining components from various methods or using these components as a launching pad for a novel technique. In \Cref{sec:notional_examples} we illustrate how our guidelines can be used for self-assessment selection and tuning in two notional examples of national interest. We conclude with a discussion of open research challenges in \Cref{sec:discussion}.

\subsection{Mathematical framework}
\label{subsec:math}

\begin{figure}[htb]
\centering
\includegraphics[width=0.8\textwidth]{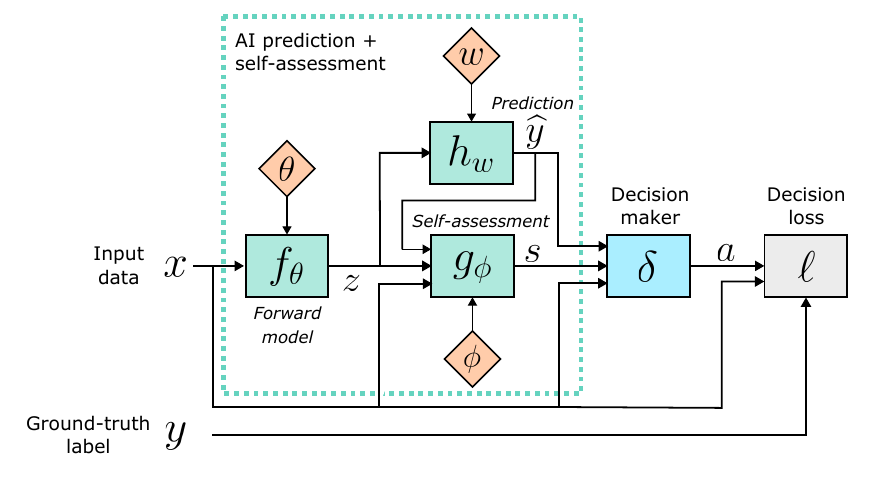}
\caption{Mathematical framework for decision-driven AI self-assessment.}
\label{fig:notation}
\end{figure}

It will sometimes be useful to describe techniques using a standardized notation and mathematical framework to discuss uncertainty-aware, decision-driven self-assessment in a unified manner. To this end, we build on notation and concepts from \textcite{kirchenbauer2022what} and \textcite{zhao2021calibrating} to develop a mathematical, decision-driven framework for general self-assessment. We depict this framework in its entirety in \Cref{fig:notation}, and describe each core component below. Although this framework is a useful tool for relating self-assessment techniques to one another and understanding them in the broader context of AI learning and decision making, \textbf{it is not necessary for a reader to understand this mathematical framework to make use of our methodology}. Therefore, the mathematical framework below should be considered optional, and readers interested in higher-level self-assessment guidance can skip to \Cref{sec:attr}.

\paragraph{AI predictive model} Let $\cX$ denote a data domain of interest (e.g., $\cX = \R^d$) with individual examples denoted by $x$. Each example $x$ is associated with a label $y \in \cY$, such as $\cY = 1 \dots k$ for classification over $k$ classes. Without loss of generality, we describe the AI model as a composition of two stages: a ``forward model'' containing the bulk of AI computation (i.e., forward pass in a neural network) and a ``prediction model'' converting the output of this computation into a specific prediction. Letting $\cZ$ denote an intermediate space to be defined shortly, we define the AI ``forward'' model as $f_\theta \colon \cX \to \cZ$, parameterized by the parameter set $\theta$ in parameter space $\Theta$. We define the secondary ``prediction'' function $h_w \colon \cZ \to \widehat{\cY}$, parameterized by $w$ in parameter space $\cW$, as a post-processing function mapping the intermediate representation $z \in \cZ$ to a specific prediction $\widehat{y} \in \widehat{\cY}$, where $\widehat{\cY}$ is a space of predicted labels. For a given example $x$, the AI model's prediction is then given by the composition $\widehat{y} = h_w(f_\theta(x))$. We separately define the intermediate representation $z \in \cZ$ and the specific model prediction $\widehat{y} \in \cY$ to make our framework as general as possible and allow for a distinction between intermediate variables produced by a model (which may be used in subsequent self-assessment) and the ultimate label prediction.

For instance, in a typical deep learning scenario, $\cZ = \Delta^{k-1}$ is the simplex of probability distributions over $k$ classes, and $f_\theta$ is a neural network with weights $\theta$ that outputs a predictive label distribution $\widehat{p} \in \Delta^{k-1}$ over $k$ classes. In the standard setting, the prediction label set $\widehat{\cY}$ is equal to the true label set $\cY$. However, this is not always the case, such as in models that can abstain from making specific predictions by allowing for an ``I don't know'' prediction $\bot$. To arrive at a single label prediction, the maximum value of $\widehat{p} = f_\theta(x)$ is taken as the predicted label, which in our framework can be described by defining the function $h = \argmax_{i} \, \widehat{p}_{i}$, where $\widehat{p}_i$ is the $i$th entry of $\widehat{p}$, and where $\cW = \varnothing$ since this predictive mapping does not require additional parameters.

\paragraph{AI self-assessment} Beyond the specific predictions $\widehat{\cY}$ produced by an AI model, we define $\cS$ as the space of outputs produced by a corresponding self-assessment technique. For example, one popular approach to self-assessment is for the model to produce a scalar probability indicating its confidence in its prediction $\widehat{y}$. In this case, $\cS = [0,1]$ is the unit interval on which this confidence value lies. To produce a self-assessment output, we define the mapping $g\colon \cX \times \cZ \times \widehat{\cY} \to \cS$, parameterized by $\phi \in \Phi$, since in the most general scenario the self-assessment might depend on the input data $x$, forward representation $z = f_\theta(x)$, and predicted output $\widehat{y} = h_w(f_\theta(x))$, as $s = g_\phi(x, f_\theta(x), h_w(f_\theta(x)))$. As an example, to describe the scalar confidence self-assessment described above, suppose that the forward model outputs $\widehat{p} = f_\theta(x)$ (where $\cZ = \Delta^{k-1}$). Then, defining $g = \max_i \, \widehat{p}_i$, we can write $s = g(f_\theta(x))$, where in this case $g$ does not depend on $x$ or $\widehat{y}$ directly (but could in general) and does not have separate parameters $\phi$ (so $\Phi = \varnothing$). Note that in some approaches where the output of $f_\theta$ relies on stochastic elements (e.g.,\ dropout), the self-assessment stage $g$ may require access to multiple forward passes to arrive at an uncertainty estimate (as in methods based on ensembles or Monte Carlo samples). Other self-assessment methods may rely only on knowledge of the forward model's parameters $\theta$, without requiring additional uncertainty parameters. For clarity we do not indicate all possible dependencies in \Cref{fig:notation} for every self-assessment scenario, and instead adopt a general framework that covers many use cases.

\paragraph{Downstream decision-making} Suppose that a decision-maker (e.g., human user, downstream software) observes the AI model's prediction $\widehat{y}$ and self-assessment $s$ to inform a downstream decision over a discrete set of possible actions $a \in \cA$. We formally define a decision-making policy $\delta\colon \cX \times \widehat{\cY} \times \cS \to \Delta^{\abs{\cA} - 1}$ that takes as input the current example $x$, AI prediction $\widehat{y}$, and self-assessment $s$ and outputs a probability distribution over the set $\cA$ of possible actions. We assume without loss of generality\footnote{This probabilistic definition still allows for the possibility of deterministic decision policies by simply defining $\delta$ as selecting a single action with probability 1.} that an action is then sampled from this probability distribution $\delta(x, \widehat{y}, s)$. Since the decision-maker is typically unaware of the true label $y$, we assume that $a \sim \delta(x, \widehat{y}, s)$ is conditionally independent of $y$, given $x$.

We define a loss function $\ell\colon \cX \times \cY \times \cA \to \R$ that measures the ``cost'' of each decision, as follows: given input example $x$ with ground-truth label $y$, if the decision-maker selects action $a$, then a cost of $\ell(x, y, a)$ is incurred; such a loss formulation is consistent with standard concepts in decision theory. Furthermore, this formulation generalizes common losses such as classification error, in which case $\cA = \cY$ and $\ell(x, y, a) = \ind[a \neq y]$. In some frameworks such as \textcite{zhao2021calibrating}, the loss function only depends on label $y$ and action $a$, but for generality, we include a dependence of $\ell$ on $x$. For a distribution $\cD_{X,Y}$ over example/label pairs $(x,y)$, we define the \emph{expected decision cost} as $C = \E_{x,y \sim \cD_{X,Y}} \, \E_{a \sim \delta(x, \widehat{y}, s)} [\ell(x, y, a)]$. Generally speaking, for a given decision policy $\delta$ and data drawn from distribution $\cD_{X,Y}$, a system designer should choose the AI forward model $f_\theta$, predictive model $h_w$, and self-assessment $g_\phi$ to jointly minimize the overall downstream decision cost $C$.

\FloatBarrier

%% file: sections/attributes.tex
\section{Self-assessment attributes and design guidelines}
\label{sec:attr}

There is a wide spectrum of approaches and techniques developed for uncertainty quantification of AI predictions; see, for example, \textcite{abdar2021review, gawlikowski2023survey}. \textbf{To categorize this varied range of techniques in a manner that aids practitioners, we organize methods according to attributes that are important factors for designing and evaluating decision-driven AI self-assessments.} In the following, we identify these key attributes and provide examples of their possible values. Some examples of how existing uncertainty estimation techniques can be characterized by their associated attributes are given in Tables~\ref{table:decision_agnostic} and~\ref{table:decision_aware}, which will be described in more detail in \Cref{sec:overview}. 

\subsection{AI task}
\label{subsec:tasks}

Uncertainty estimation is focused on characterizing the uncertainty of the AI model’s output/prediction given an input. Hence, the relevant representation and estimation mechanisms for uncertainty will naturally depend on the underlying AI task. It is also important to consider the desired performance for the underlying AI tasks, as applying some uncertainty estimation techniques may impact the AI’s performance for its given task. The majority of existing research on uncertainty estimation has been focused on classification and regression, but recent efforts have started examining a broader range of AI tasks:
\begin{itemize}
    \item Classification with performance measures based on the prediction confusion matrix, rather than average accuracy
    \item Object detection
    \item Regression
    \item State estimation (tracking)
    \item Segmentation
    \item Reinforcement learning
    \item Generative tasks
\end{itemize}
Although uncertainty quantification techniques have been developed for several of these tasks (e.g., regression \cite{dheur2023large, dheur2024probabilistic, bui2024density, capone2023sharp}, object detection \cite{popordanoska2024beyond, oksuz2023towards, munir2023bridging, pathiraja2023multiclass}), here we mostly focus on classification problems due to their ubiquity across a variety of practical AI problem settings. It would be straightforward to apply our framework more broadly to various other AI tasks, and we leave to future work an in-depth study of decision-driven self-assessment for non-classification AI tasks such as those listed here.

\subsection{Uncertainty representation}
An important feature of an uncertainty estimation technique is which notion of uncertainty is estimated and how it is represented. Different notions of uncertainty will naturally call for different representations and technical approaches. We identify the most common types of representations below: 
\begin{itemize}
\item 
\textbf{Scalar confidence}: often interpreted as an estimate of the posterior probability of the model’s decision, given the observations.
\item 
\textbf{Confidence interval/set}: compact set in the output space with a probability bound (often referred to as the “coverage” for confidence intervals); could be an interval for continuous output or a discrete set for categorical output.
\item
\textbf{Parametric density/distribution}: An approximation to the output distribution, often through approximation with a Gaussian distribution (first and second moments).
\item
\textbf{Out-of-distribution (OOD) score/probability}: A measure of likelihood that an input to the model ``comes from'' a distribution distinct from the training data distribution. OOD is generally ill-defined without further assumptions on the underlying distributional shifts \cite{garg2022leveraging, david2010impossibility, lipton2018detecting}. Conceptually, a high OOD score for an input implies that the model trained and calibrated with training data provides limited information on the ``true'' prediction for the input. 
\end{itemize}

\subsection{Generic metrics}
\label{subsec:generic}
Given a representation of AI uncertainty, there are ``generic'' metrics that measure the quality of the underlying estimation problem. We refer to these as “generic” since they often do not consider specific impacts on downstream decisions. In selecting and tuning decision-driven self-assessment techniques, these generic metrics may not be the appropriate metric to optimize directly without further accounting for downstream decision-making. However, these metrics are often central to designing analogous metrics that directly take downstream decision costs into consideration.

\begin{itemize}
    \item 
    \textbf{Metrics for scalar confidence}: Expected calibration error (ECE) \cite{naeini2015obtaining}, maximum calibration error (MCE) \cite{naeini2015obtaining}, negative log likelihood (NLL), Brier score, class-wise ECE, adaptive calibration error (ACE) \cite{nixon2019measuring}. Such metrics for calibration error have been generalized in a unified metric known as Generalized Expected Calibration Error (GECE) \cite{kirchenbauer2022what}.
    \item
    \textbf{Metrics for confidence interval/set}: Prediction Interval Coverage Probability (PICP) and Mean Prediction Interval Width (MPIW) are the standard metrics to measure the associated accuracy and “tightness/specificity” of confidence intervals. We can generalize their definitions for discrete confidence sets by defining an appropriate notion for the “size” of a set (e.g., by cardinality).
    \item 
    \textbf{Metrics for parametric distribution}: Several metrics can be used to measure the quality of the distributional approximation including negative log likelihood and an estimation of the Kullback-Leibler or related notions of divergence.
    \item 
    \textbf{Metrics for OOD}: The standard approach to evaluate OOD performance is to view the problem as a binary classification and resort to metrics such as the area under the receiver operating characteristic (AUROC) and the area under the precision-recall curve (AUPRC). It is important to characterize the anticipated distribution shifts (e.g., covariate versus concept shifts) OOD is intended to address to interpret these metrics accordingly (see \cite{yang2024imagenetood}). 
\end{itemize}

\subsection{Estimation mechanisms}
Uncertainty estimation can be broadly categorized into the following three mechanisms, differentiated by when they occur in training, and how much additional modeling is required beyond the native model $f_\theta$:
\begin{itemize}
    \item \textbf{Post-hoc techniques}: These techniques are applied to a trained ML model without further tuning the forward model weights $\theta$. Instead, post-hoc adjustments are made to the parameters $\phi$ determining the behavior of the self-assessment stage $g_\phi$ itself.
    \item \textbf{Integral to training}: These techniques require access to the training process of the AI model, i.e., the training algorithm for $f_\theta$. For instance, MC dropout \cite{gal2016dropout} utilizes the same dropout mechanism used during training to sample from a posterior distribution of network weights, rather than requiring a separate re-training step. Similarly, ensemble methods \cite{lakshminarayanan2017simple} rely on training multiple candidate models, and the disagreement between such models captures a notion of uncertainty.
    \item \textbf{Intrinsic to model}: rather than introducing self-assessment as an auxiliary modification for an AI model originally designed for prediction-only, there exists a class of self-assessment approaches where the elements of uncertainty quantification are \emph{intrinsic} to the predictive model itself. For instance, Bayesian Neural Networks explicitly maintain a probability distribution over network weights. Typically, self-assessment techniques that are ``intrinsic to the model'' involve fundamental changes to the model itself, such as significant architectural modifications (e.g., outputting the mean and variance of a distribution rather than a pointwise prediction).
    
\end{itemize}
It is important to consider the underlying mechanism when considering an uncertainty estimation approach as the AI/ML model may have been trained a priori and prevent the applications of approaches requiring fundamental changes to training and/or model architecture selection. In addition, approaches intrinsic to the model will likely impact the performance of the underlying ML task.    

\subsection{Design parameters}

Given the variety of metrics that might be used to assess the performance of any particular self-assessment technique, it is natural to utilize these metrics to guide the selection of which self-assessment technique to deploy and how to set any relevant \emph{SeA design parameters} governing self-assessment behavior. As discussed in \Cref{subsec:math} and \Cref{fig:notation}, when possible we distinguish between the parameters $\theta$ involved in the AI ``forward'' model making label predictions, and any SeA design parameters $\phi$ involved in tuning the self-assessment model itself.\footnote{As mentioned in \Cref{subsec:math}, it may not always be possible to make this distinction, especially in SeA approaches that are intrinsic to the AI model or integral to training. For example, the dropout probability $p$ in Monte Carlo Dropout \cite{gal2016dropout} affects both regularization during forward model training, and the label distribution obtained during self-assessment. In such cases, we might still consider such shared parameters between AI prediction and self-assessment to be ``tunable'' by a practitioner.} While the relevant SeA design parameters depend on the specific technique being utilized, a few representative examples include:
\begin{itemize}
\item Temperature $T > 0$ controlling the ``sharpness'' of an estimated label distribution \cite{guo2017on}. In the notation of our framework let $\cZ = \R^k$, define $z = f_\theta(x)$ as the ``logit'' vector output by a forward model, define $\cS = \Delta^{k-1}$, $\phi = \{T\}$, and let $\sigma(\cdot)$ notate the softmax operation. We can then ``scale'' an AI model's output label distribution by letting $g_\phi(z) = \sigma(\nicefrac{z}{T})$. Depending on the value of $T$, this operation softens or sharpens an estimated label distribution from the forward model to ensure that the scaled distribution is well-calibrated or satisfies other properties.
\item Error rate $0 < \alpha < 1$ utilized in forming a conformal set \cite{angelopoulos2023conformal}. In our framework, let $\cZ = \R^k$ and $z = f_\theta(x)$ denote a score vector over $k$ classes, $\phi = \{\alpha\}$, and let $\cS = P(\widehat{\cY})$ be the power set (set of all subsets) of $\widehat{\cY}$. Then $g_\phi(x) = \{y' : f_\theta(x)[y'] \ge \tau\}$ where $f_\theta(x)[y']$ is the entry of $f_\theta(x)$ corresponding to label $y'$, and $\tau$ is computed such that on average over a held-out calibration set $g_\phi(x)$ contains the true label $y$ with probability at least $1 - \alpha$.
\end{itemize}

In certain settings, such SeA design parameters might be numerically optimized by an algorithm minimizing a relevant loss function (e.g., optimizing temperature $T$ to minimize negative log-likelihood \cite{guo2017on}). In other cases, a practitioner might set such parameters manually by monitoring their effect on downstream evaluation metrics of self-assessment performance, or may select parameters based on desired SeA characteristics (e.g.,\ error rate $\alpha$ in conformal prediction). Here, we take a broad perspective and loosely refer to the process of either numerically adjusting or hand-tuning SeA design parameters as ``optimization.'' Furthermore, in typical machine learning nomenclature one would distinguish between parameters that directly tune SeA model behavior and any \emph{hyperparameters} associated with SeA design, which can also be tuned or selected from to affect self-assessment performance. Examples of such hyperparameters might include:
\begin{itemize}
\item Binning scheme (number of bins, bin edges) utilized in the histogram-based calibration of estimated label distributions \cite{naeini2015obtaining}.
\item Tuning regularization parameters in a loss function for training $g_\theta$, including choosing a different loss function altogether. Deciding on a different loss function can affect, for instance, whether or not a self-assessment technique is risk-aware.
\item Costs/weights on different prediction errors or levels of self-assessment confidence (e.g., overconfidence can be costly since a human decision-maker may blindly trust the results \cite{corvelo2023human}).
\item Selection of calibration set used for tuning self-assessment parameters.
\item Choice of models used in an ensemble and how uncertainty is estimated from this ensemble.
\end{itemize}
In a slight abuse of terminology, for conciseness we continue to refer to such hyparameters as simply being other SeA ``design parameters'' that a practitioner might tune or optimize. What all of the ``parameters'' described in this section have in common is that they are all intrinsic to the self-assessment technique, and need to be set by some means.

\subsection{Downstream decisions}
Even though not an attribute of an estimation technique, the downstream decision-maker that will “consume” the output from the AI model and its self-assessment shall dictate the applicable uncertainty representations. For example, a human user may not be able to interpret a complex label distribution to arrive at an informed decision, and it may be better to provide an uncalibrated yet interpretable distribution instead. Beyond compatibility with the decision-maker, the output of self-assessment mapping $g_\theta$ may drastically affect which downstream decisions are made, even for non-human decision-makers. For example, a downstream Kalman-filter based tracking algorithm may assume a covariance matrix (to characterize observation uncertainty) associated with each detection to combine detections from multiple sensors. If this covariance matrix is provided by the self-assessment model $g_\theta$, then $g$, in effect, significantly impacts the downstream performance of the tracking algorithm.

While many self-assessment techniques are designed to increase decision-maker trust in AI predictions and effectively convey uncertainty, it is not always the case that explicit effects on downstream decisions and associated costs are considered during self-assessment design and optimization. Many self-assessment techniques are derived, optimized, and evaluated based on a set of ``generic'' statistical metrics measuring various properties of their uncertainty outputs (e.g., distribution calibration, see \Cref{subsec:generic}), which are usually calculated independently from subsequent downstream decisions; here, we refer to such methods as being \emph{decision-agnostic}. This stands in contrast to self-assessment techniques that \emph{explicitly} consider self-assessment outputs' effects on downstream decisions, which we categorize as being \emph{decision-aware}. For example, as described in \Cref{subsec:decision_aware},  certain self-assessment approaches optimize SeA parameters to minimize the expected downstream decision cost explicitly.

\subsection{Guided self-assessment design}
\label{subsec:guided}

\begin{figure}[htb]
\centering
\includegraphics[width=0.6\textwidth]{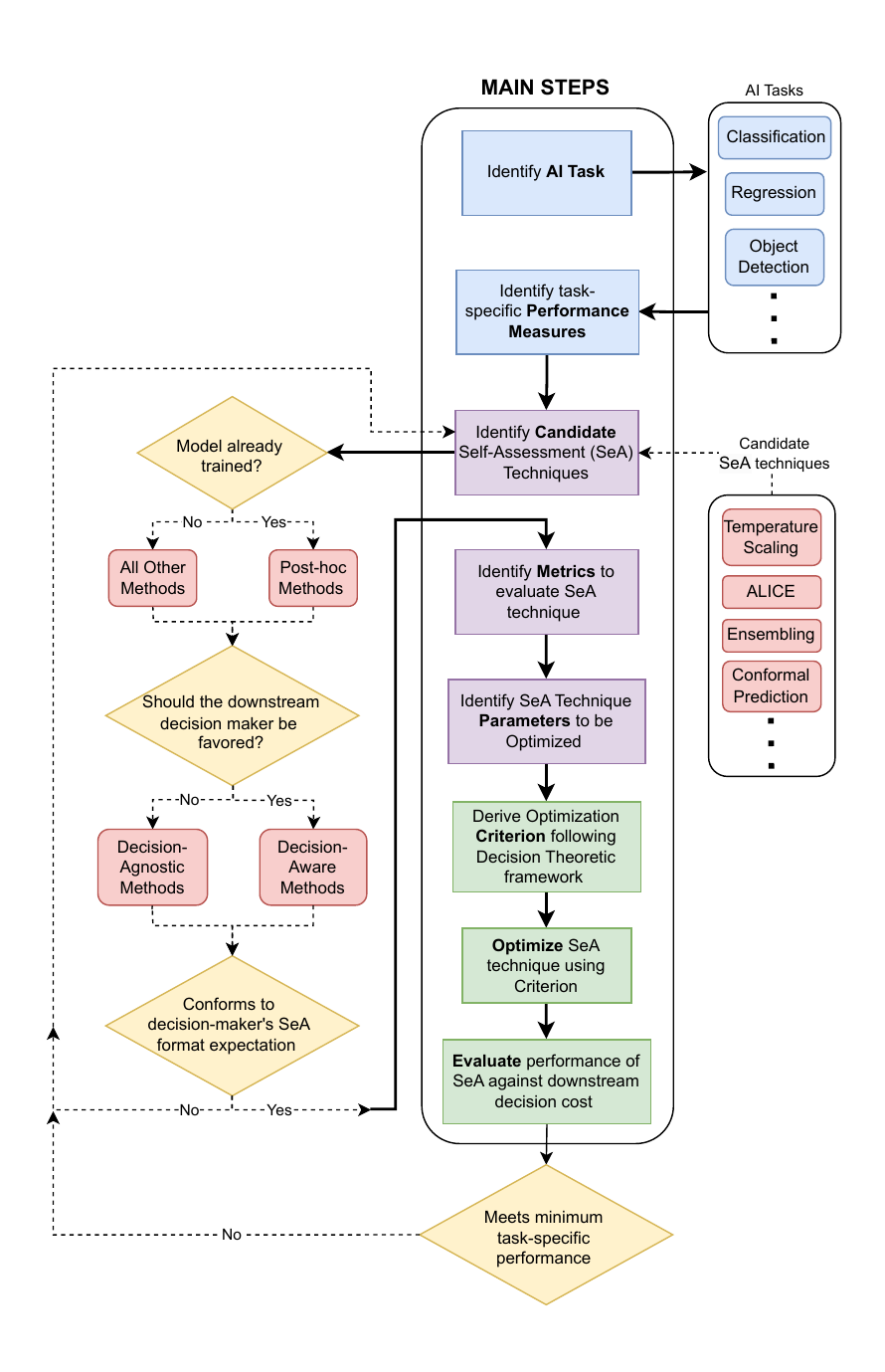}
\caption{Methodology for a practitioner to downselect and optimize candidate self-assessment techniques for their application at hand. For candidate self-assessment techniques, refer to \Cref{sec:overview} along with \Cref{table:decision_agnostic,table:decision_aware}. For illustrative examples utilizing this methodology, refer to \Cref{sec:notional_examples}.}
\label{fig:flowchart}
\end{figure}

The key attributes outlined above define the important ``dimensions'' to consider when selecting, tuning, and evaluating uncertainty estimation for AI self-assessment. As these dimensions are not necessarily orthogonal, it is unlikely that a monolithic decision process (i.e., a single decision tree) exists for all use cases to guide a user towards an optimal self-assessment technique. Here, we postulate a likely ``decision flow'' (depicted in \Cref{fig:flowchart}) to illustrate a typical self-assessment design process:

\begin{enumerate}[parsep=0pt]
    \item Identify candidate uncertainty estimation approaches based on the \textbf{AI task} at hand;
    \begin{enumerate}
        \item If a model has been trained, then only the \textbf{post-hoc} approaches are applicable.
        \item Determine if an approach could have a significant negative impact on the associated task-specific \textbf{performance measure}.
    \end{enumerate}
    \item Determine the appropriate \textbf{uncertainty representation} taking into account the model (if already trained) and the \textbf{downstream decision} (what the downstream decision-maker can consume).
    \item Examine the published benchmark on the \textbf{generic metrics} associated with the uncertainty representation to develop a qualitative analysis of candidate approaches.
    \item Identify tunable \textbf{optimization parameters} for the candidate approaches and derive an optimization criterion following our decision-theoretic framework.
    \item Optimize uncertainty estimation against the optimization criterion and evaluate the performance based on the expected cost associated with the \textbf{downstream decision}. 
\end{enumerate}

\FloatBarrier

%% file: sections/overview.tex
\section{Overview of self-assessment techniques}
\label{sec:overview}  

By following the guidelines presented in \Cref{subsec:guided} and in \Cref{fig:flowchart}, a user can arrive at a method for self-assessment appropriate for their particular problem scenario. In this section, we provide an overview of relevant self-assessment techniques and describe them according to the attributes presented in \Cref{sec:attr}. The set of techniques considered here should be considered a representative, but not necessarily exhaustive, suite of complementary approaches for self-assessment. As new approaches are developed, they can easily be categorized along the dimensions of \Cref{sec:attr} and added here. In \Cref{subsec:decision_agnostic}, we detail various approaches to \emph{decision-agnostic} self-assessment (summarized in \Cref{table:decision_agnostic}), followed by a representative suite of \emph{decision-aware} techniques in \Cref{subsec:decision_aware} (summarized in \Cref{table:decision_aware}).

\subsection{Decision-agnostic self-assessment}
\label{subsec:decision_agnostic}

In this section we present a representative overview of popular and distinctive methods for decision-agnostic self-assessment (summarized in \Cref{table:decision_agnostic}). First, we outline approaches rooted in post-hoc adjustment of a model's class label distribution to minimize various notions of calibration error. Then, we pivot to discuss methods with uncertainty representations intrinsic to the model or integral to the model's training process. Finally, we briefly mention several techniques with noteworthy uncertainty representations distinct from direct estimates of label or model uncertainty.

\paragraph{Post-hoc approaches} One of the most common approaches to self-assessment is to output the estimated probability of a model's prediction as a scalar notion of ``confidence'' by letting $h(z) = \argmax_{y'} \, z_{y'}$ and $g(z) = \max_{y'} \, \sigma(z)[y']$, where $z$ is a vector of logits output by an AI forward model, $\sigma(z) = \operatorname{softmax}(z)$, and $\sigma(z)[y']$ denotes the softmax probability corresponding to class $y'$. However, it has generally been established that such a confidence measure is not necessarily well-calibrated, especially for modern neural networks \cite{guo2017on}. A simple and practical strategy to remedy this miscalibration is to apply post-hoc calibration by sharpening or smoothing the estimated confidence scores with a temperature parameter $T > 0$ by letting $g_T(z) = \max_{y'} \, \sigma(\nicefrac{z}{T})[y']$. For small values of $T$, the model places higher confidence in its prediction, and for larger temperature values, the output probabilities are ``softened,'' resulting in higher model uncertainty. Such \emph{temperature scaling} for uncertainty quantification was introduced in \textcite{guo2017on} and extended to other scalar notions of confidence by \textcite{yona2022useful}.

Building on this approach, \textcite{guo2017on} introduced an extension to temperature scaling where a linear transformation $\bW z + \bb$ is applied to the logits $z$ before the softmax operator in lieu of a simple scaling by $\nicefrac{1}{T}$. \textcite{kull2019beyond} build on this concept even further by applying such a transform to \emph{log probabilities} before subsequently applying softmax, which was shown to be equivalent to learning a Dirichlet distribution over the class labels as a means of calibration. Rather than using a parametric transformation such as temperature or matrix scaling to calibrate an output distribution, \textcite{zadrozny2001obtaining} introduced a non-parametric approach to post-hoc calibration known as \emph{histogram binning}. In this approach (applicable to binary classification with $k=2$), data points are first binned according to their sorted predicted probabilities, and then each point's positive class probability is estimated by computing the fraction of positive instances empirically observed in the bin (theoretical guarantees were shown for histogram binning in \textcite{gupta2021distribution}). \textcite{naeini2015obtaining} extend this technique by simultaneously employing multiple binning schemes in a Bayesian model to address the limitations of histogram binning associated with having fixed bin edges. We note that principles from scaling-based methods and histogram binning have been combined in a scaling-binning calibrator that enjoys theoretical guarantees on calibration error \cite{kumar2019verified}.

\paragraph{Integral to training / intrinsic to model} Beyond self-assessment methods that applying a post-hoc re-calibration method to model outputs, as discussed in \Cref{sec:attr} there exist other approaches tied to model training itself, which sometimes include uncertainty representation components intrinsic to the forward model. Such approaches vary in their degree of explicitness in how uncertainty is represented. At one extreme, deep ensembles \cite{lakshminarayanan2017simple} train separate networks with varying initializations, and average their output predictions to implicitly account for model uncertainty. Similarly in \emph{MC Dropout}, uncertainty is represented implicitly by making multiple stochastic forward passess through a model trained with dropout, which has been shown to approximately solve a variational inference problem \cite{gal2016dropout}. 

At the other end of the extreme, \textcite{blundell2015weight} directly model a posterior distribution over network weights via a variational Gaussian approximation, explicitly maintaining a mean and variance for each network weight. Training such variational inference approaches can either rely on Monte Carlo sampling as in \textcite{blundell2015weight} or can instead leverage analytic evidence lower bound evaluation \cite{wright2024analytic}. Stochastic Weight Averaging-Gaussian (SWAG) takes a similar approach by maintaining a Gaussian posterior distribution for model weights that includes low-rank elements \cite{maddox2019simple}. Epinets \cite{osband2023epistemic} utilize a sampling scheme as in \textcite{lakshminarayanan2017simple} to maintain an ensemble of models, but also introduce a new architecture to supplement existing base networks that separates out deterministic network components from stochastic components (\textcite{kendall2017what} also use a network modification to model both aleatoric and epistemic uncertainties). Unlike epinets and Bayesian Neural Networks, \textcite{kumar2018trainable} do not modify the network's architecture and simply output a conditional label distribution over each class. However, they introduce Maximum Mean Calibration Error (MMCE) as a means of regularizing network training to learn a calibrated output distribution.

\paragraph{Distinctive uncertainty representations} There exists a wide variety of self-assessment techniques with uncertainty representations that are somewhat distinctive from the approaches described so far in this section. One major category of such methods is those techniques performing \emph{out-of-distribution} (OOD) detection, to determine whether a given input point $x$ lies ``within'' the data distribution of points that a model is expected to generalize on. Detecting out-of-distribution points can potentially serve as a valuable form of self-assessment by hedging a model's prediction with the fact that the point is ``unlike'' the training data. Within OOD detection, there are various types of data distribution shifts that might constitute a point being ``out of distribution,'' including both semantic and covariante shifts \cite{yang2024imagenetood}. It has recently been argued in \textcite{yang2024imagenetood} that modern OOD approaches may not perform as expected under these various types of shifts, and instead the authors provide empirical support for a simple baseline method known as Maximum Softmax Probability (MSP) \cite{hendrycks2017a} which takes $g(z) = \ind[\max_{y'} \sigma(z)[y'] > \tau]$ as an indicator of a point's In or Out of distribution status. There are a wide variety of other approaches for OOD detection (see \textcite{yang2024generalized} for a comprehensive survey), including score-based methods proposing alternate scoring functions besides MSP to measure a point's In/Out status (such as energy-based methods \cite{liu2020energy}).

To enable informative evaluation and fair comparison of OOD detection techniques, it is important that we recognize whether a method makes an explicit assumption on the nature of the distribution shift it was designed to detect. For example, techniques developed specifically for label shift such as \cite{lipton2018detecting} are not expected to perform well for detection of covariate shift. Other techniques developed without explicit assumptions on distributional shifts such as MSP may have disparate performance across different shifts.   Such an understanding is critical for identifying candidate approaches that are likely to succeed in addressing the distributional shifts anticipated for each application.

A self-assessment approach similar in spirit to OOD detection is \emph{learning to reject}, where a model not only makes label predictions but learns when to abstain from making predictions on points it is unsure about \cite{cortes2016learning}, mitigating the risk of providing a downstream decision-maker with an untrustworthy prediction (specifically, abstaining on those potentially erroneous predictions that might incur a large cost if they lead to poor downstream decisions). Learning to reject is a self-assessment type worthy of consideration when interfacing with a downstream decision-maker, and we refer interested readers to the surveys in \cite{yang2024generalized, zhang2023survey} for additional details and references.

There have also been related concepts proposed in the literature such as \emph{atypicality} estimation, which for a given point provides a measure of how likely such a point is to occur in a training distribution, rather than explicitly detecting In/Out status. \textcite{yuksekgonul2023beyond} utilize this concept in developing an atypicality-driven post-hoc re-calibration method (AAR). \textcite{rajendran2019accurate} combine several concepts such as label distribution calibration with OOD detection to derive a post-hoc self-assessment score known as ALICE, providing a conservative estimate for a model's confidence that takes into account both label distribution estimation and distribution shifts. Furthermore, ALICE does not simply re-calibrate a distribution or detect points as In/Out: instead, it combines these concepts into a pointwise notion of \emph{competence} estimating the probability that a model's error is below a specified threshold.

Another major approach utilizing an entirely distinct uncertainty representation is the class of methods known by \emph{conformal prediction} \cite{angelopoulos2023conformal}. In conformal prediction, the self-assessment model does not output any direct probability estimates (either with respect to the input distribution or label distribution) to a downstream decision-maker. Instead, uncertainty is conveyed \emph{implicitly} to the decision maker by expanding the set of predictions returned beyond a single prediction $\widehat{y}$. Here, a set of predictions $\mathcal{C}$ is returned that is selected to contain the true label $y$ with probability $1 - \alpha$ for a pre-specified value of $\alpha$. In fact, this approach enjoys rigorous theoretical guarantees bounding the probability that $\mathcal{C}$ contains $y$ towards $1 - \alpha$ (such guarantees are made in a marginal sense; see \textcite{angelopoulos2023conformal} for discussion). Conformal prediction sets have the benefit of potentially being more interpretable to downstream decision-makers, since they make statements about inclusion or exclusion in the prediction set $\mathcal{C}$ rather than ascribing numerical notions of probability, which may be difficult to interpret by human decision-makers.

\begin{table}[htb]
\footnotesize
\centering
\input{figures/decision_agnostic_table}
\caption{Overview of decision-agnostic self-assessment techniques.}
\label{table:decision_agnostic}
\end{table}

\FloatBarrier

\subsection{Decision-aware self-assessment}
\label{subsec:decision_aware}

Qualitatively, the mere presence of AI self-assessment may encourage downstream decision-makers about the trustworthiness of the AI's predictions. However, to quantitatively benefit downstream decision processes, the self-assessment must directly improve metrics measuring downstream decision quality. In the context of our mathematical framework, this is achieved by a self-assessment strategy being designed and tuned to optimize the expected downstream decision cost $C$ with respect to a cost function $\ell(x, y, a)$ that is appropriate for a user's particular application (see \Cref{sec:intro} and \Cref{fig:notation}).

Choosing self-assessment strategies that optimize or at least account for such downstream decision costs goes a step beyond optimizing ``generic metrics'' for self-assessment that may not result in optimal downstream decisions. For example, \emph{uncalibrated} label distributions taking into account the nuances of human psychology can sometimes result in more optimal human decision-maker performance than by using calibrated distributions that don't explicitly take the decision-maker into account. In this vein, there is a significant body of human factors research analyzing the joint performance of human-AI decision-making systems under various types of self-assessment \cite{corvelo2023human, babbar2022on, cresswell2024conformal}. In recent years, a suite of self-assessment techniques has emerged that are explicitly designed to take these factors into account and optimize downstream decision costs; we highlight a representative set of such methods here (summarized in \Cref{table:decision_aware}).

\textcite{marx2023calibration} develop a general, kernel-based calibration method that is integral to model training, where different selections of kernels result in certain types of calibration being optimized. \textcite{kumar2018trainable} can be seen as a special case of this more general framework. As an example, a kernel based on downstream decision losses is presented as a special case of their more general framework, and evaluated on a threshold-based decision-making task. This approach strives to minimize Decision Calibration Error (DCE) which measures the difference between the expected decision cost as calculated from the self-assessment distribution, versus the true expected decision cost. By ensuring a low DCE, downstream-decision makers can trust that estimates of decision cost calculated from the self-assessment output are reliable estimates of the true cost for each candidate decision, to better inform decision-making.

This work builds on the earlier work in \textcite{zhao2021calibrating}, which first introduced the notion of decision calibration and devised an algorithm to re-calibrate a given label distribution $\widehat{p}$ to minimize decision calibration error. However, this earlier work is calibrated to the set of decision losses over a fixed number of actions rather than being calibrated with respect to a specific decision loss $\ell$, and is only applicable to Bayes optimal decision makers $\delta$, which may not be applicable in real-world scenarios. \textcite{sahoo2021reliable} address the setting of thresholded decisions on a regression model, and present an algorithm for re-calibrating a probability distribution over regression values to minimize the difference between estimated and true losses incurred under a threshold decision model.

Although \textcite{marx2023calibration, zhao2021calibrating} take into account downstream decision costs in their self-assessment tuning, they only allow for a Bayes optimal decision maker $\delta$, which is likely to be unrealistic in practical settings. Instead, \textcite{vodrahalli2022uncalibrated} address this design aspect explicitly by \emph{modeling} the decision-maker $\delta$ using a separate predictive model. Once $\delta$ is estimated from human behavior training data, along the lines of temperature scaling the AI self-assessment (in this context, called the AI ``advice'') notated by $A$ is re-calibrated with parameters $\alpha, \beta \ge 0$ to arrive at a re-calibrated scalar confidence value $g(A) = 1 / (1 + \exp(-\operatorname{sign}(A)(\alpha \abs{A} + \beta))).$ These constants are selected to minimize the binary cross-entropy loss of the human decision-maker's (modeled via $\delta$) ultimate decision, after viewing the re-calibrated AI ``advice.'' While presenting a promising concept, the author's also note that this framework is ``not currently suitable for practical use.''

In a similar vein, \cite{straitouri2023improving} uses an estimate of the decision-maker's confusion matrix over classes $\widehat{Y}$ with respect to the ground-truth class $y$ to choose an optimal coverage probability for a conformal set to minimize the decision-maker's ultimate probability of decision error. \textcite{kerrigan2021combining} similarly utilize confusion matrix estimates to combine human and AI decisions into joint decisions that outperform either the AI or human making decisions on their own. \textcite{bansal2021is} expand the considered downstream loss function $\ell$ to account for unequal costs of correct or incorrect decisions. The authors utilize these cost parameters along with several decision-theoretic assumptions to derive a novel loss function equal to the expected utility, which is used directly for model training.

In contrast to many of the decision-agnostic methods described in the previous section, what unifies the decision-aware approaches presented here is their \emph{explicit} consideration of the downstream decision maker in designing an appropriate SeA method, taking the resulting decision costs into account. This class of methods holds promise as an avenue to ensure that the joint human-AI \emph{system} is performing optimally in a given problem scenario. Explicitly incorporating downstream decision costs into the selection and optimization of self-assessment techniques is a relatively nascent area of research, and there are still many open directions in designing self-assessment strategies that truly bridge the gap between decision theory and practical applications (see \Cref{sec:discussion} for a discussion of potential research avenues). 

\begin{table}[htb]
\footnotesize
\centering
\input{figures/decision_aware_table}
\caption{Overview of decision-aware self-assessment techniques.}
\label{table:decision_aware}
\end{table}

\FloatBarrier

%% file: figures/decision_agnostic_table.tex
\makebox[\textwidth][c]{\resizebox{1.2\textwidth}{!}{%
\begin{tabular}{ | m{3cm} | m{2cm} | m{4cm}| m{4cm} | m{5cm} | m{4cm} |}
  \hline
  \textbf{Technique} & \textbf{AI task} & \textbf{Uncertainty representation} & \textbf{Generic metrics} & \textbf{Estimation mechanism} & \textbf{Tuning design parameters} \\ 
  \hhline{|=|=|=|=|=|=|}
  \multicolumn{6}{|l|}{\textbf{Post-hoc methods}}\\
  \hline
  Temperature scaling \cite{guo2017on} & Multi-class classification & Scalar confidence (prediction probability) & ECE, Brier score, NLL & Post-hoc minimization of NLL (requires calibration set) & Temperature $T > 0$\\
  \hline
  Temperature scaling with alternate confidence measures \cite{yona2022useful} & Multi-class classification & Scalar confidence options: maximum probability, difference of top-2 probabilities, weighted difference of top-3 probabilities, label entropy & $\ell_2$ calibration error, sharpness (variation in error across confidence values), ECE, ACE, evaluated with fixed and adaptive binning & Post-hoc minimization of $\ell_2$ calibration error (requires calibration set) & Temperature $T > 0$\\
  \hline
  Dirichlet calibration \cite{kull2019beyond} & Multi-class classification & Label distribution & Accuracy, log-loss, Brier score, MCE, ECE, classwise-ECE, Significance measures & Post-hoc (calibration set required) minimization of log-loss with ODIR regularizer & Linear weights and bias $\bW$, $\bb$; ODIR regularization hyperparameters $\lambda, \mu$\\
  \hline
  Bayesian Binning into Quantiles (BBQ) \cite{naeini2015obtaining} & Binary classification & Scalar probability & Discrimination power (Accuracy, area under ROC curve), Calibration error (RMSE, ECE, MCE) & Non-parametric calibration: post-hoc binning model averaging over possible binning models & Binning model: number of bins, partitioning of predictions into bins, Beta distribution parameters (per bin)\\
  \hline
  \hhline{|=|=|=|=|=|=|}
  \multicolumn{6}{|l|}{\textbf{Integral to training / intrinsic to model}}\\
  \hline
  Deep ensembles \cite{lakshminarayanan2017simple} & Multi-class classification, regression & Output distribution $p(y \mid x)$ averaged over mixture & NLL, RMSE (regression), classification error, Brier Score (classification) & Independently train each network from random initialization with adversarial training (integral to training, intrinsic to model) & Proper scoring rule $\ell$, Number of ensemble networks $M$, Adversarial training perturbation size $\epsilon$\\
  \hline
  Monte Carlo Dropout \cite{gal2016dropout} & Multi-class classification, regression, reinforcement learning & Label distribution & RMSE, log likelihood & Train model with dropout, sample from distribution with stochastic forward iterations & Dropout probability $p$, number of forward iterations $T$\\
  \hline
  Bayes by Backprop \cite{blundell2015weight} & Multi-class classification, regression, contextual bandits & Posterior distribution over network weights (diagonal Gaussian) & Variational free energy & Gradient descent with unbiased Monte Carlo gradients (integral to training, intrinsic to model) & Gaussian mean $\mu$ and variance $\sigma$ vectors, prior parameters $\pi$, $\sigma_1$, $\sigma_2$\\
  \hline
  SWAG \cite{maddox2019simple} & Multi-class classification, regression & Gaussian distribution over model weights, average label distribution over models & NLL, confidence-accuracy difference & Average first and second moments over SGD iterates & Gaussian mean $\theta_{\mathrm{SWA}}$, Gaussian diagonal covariance $\Sigma_{\mathrm{diag}}$ and low-rank covariance columns $\widehat{D}$\\
  \hline
  Epinet (Epistemic Neural Networks) \cite{osband2023epistemic} & Multi-class classification & Reference distribution over epistemic indices, stochastic epinet addition to model output & Classification error, marginal log-loss, joint log-loss & Log loss with $\ell_2$ regularizer minimized over base network (non-stochastic) + epinet (stochastic) & $\ell_2$ hyperparameter $\lambda$, prior network, index dimension\\
  \hline
  Maximum Mean Calibration Error (MMCE) \cite{kumar2018trainable} & Multi-class classification & Scalar confidence & ECE, Brier score, NLL & Optimize model parameters over NLL + MMCE loss (integral to training, intrinsic to model) & MMCE penalty $\lambda$, kernel $k(\cdot, \cdot)$\\
  \hhline{|=|=|=|=|=|=|}
  \multicolumn{6}{|l|}{\textbf{Distinctive uncertainty representations}}\\
  \hline
  Thresholding-based OOD detection (e.g.,\ maximum softmax probability) \cite{hendrycks2017a, yang2024imagenetood, yang2024generalized} & Multi-class classification & In-distribution (ID) scoring & AUROC/AUPR for error detection, In/Out detection, new class detection & Threshold scoring function $g(z)$ (e.g., $g(z) = \ind[\max_{y'} \sigma(z)[y'] > \tau]$ as in MSP \cite{hendrycks2017a}) to determine In/Out status & ID scoring function, ID scoring threshold $\tau$\\
  \hline
  Atypicality-Aware Recalibration (AAR) \cite{yuksekgonul2023beyond} & Multi-class classification & Pointwise atypicality estimates: input atypicality, class atypicality & Classification accuracy, ECE & Post-hoc fitting of penultimate GMM from training data (maximum likelihood), minimize cross-entropy loss of AAR distribution over calibration set & Penultimate layer Gaussian mixture means $\widehat{\mu}_c$ and shared covariance $\widehat{\Sigma}$, Class-wise tunable scores, Quadratic coefficients $c_2,c_1,c_0$\\
  \hline
  Accurate layerwise interpretable competency estimation (ALICE) \cite{rajendran2019accurate} & Multi-class classification & Pointwise competence & mean Average Precision (mAP) & Competence estimator calculated by combining calibrated label distribution with distance-based OOD measure & Error threshold $\delta$, choice of error function $\mathcal{E}$\\
  \hline
  Conformal prediction \cite{angelopoulos2023conformal} & Supervised learning (general) & Prediction set over outputs & PICP / MPIW & Score ranking over calibration set, quantile thresholding & Score function $s(x,y)$, coverage error rate $\alpha$\\
  \hline
\end{tabular}}}

%% file: figures/decision_aware_table.tex
\makebox[\textwidth][c]{\resizebox{1.2\textwidth}{!}{%
\begin{tabular}{ | m{3cm} | m{2cm} | m{4cm}| m{4cm} | m{5cm} | m{4cm} |}
  \hline
  \textbf{Technique} & \textbf{AI task} & \textbf{Uncertainty representation} & \textbf{Generic metrics} & \textbf{Estimation mechanism} & \textbf{Tuning design parameters} \\ 
  \hhline{|=|=|=|=|=|=|}
  Maximum mean discrepancy \cite{marx2023calibration} & Multi-class classification, regression & Label distribution & Decision Calibration Error (DCE): $\ell_2$ error of expected decision loss & Training time minimization of NLL + Maximum Mean Discrepancy (MMD) (no post-hoc calibration set) & MMD tradeoff parameter $\lambda$, choice of kernel $k(\cdot, \cdot)$\\
  \hline
  Decision calibration \cite{zhao2021calibrating} & Multi-class classification & Label distribution & Decision calibration error & Post-hoc re-calibration of initial label distribution (requires calibration set). Non-specific decision loss $\ell$ (calibrates to all loss functions over fixed number of actions), limited to Bayes optimal decision-makers & Decision-calibration error tolerance $\eps$\\
  \hline
  Threshold calibration \cite{sahoo2021reliable} & Regression, binary action space (based on thresholded regression value) & Cumulative distribution function over regression values (``forecaster'') & Reliability gap (estimated vs.\ true decision cost), true decision cost incurred & Given uncalibrated forecaster, sequentially re-calibrate using isotonic regression (requires validation set) until threshold calibration error (TCE) minimized & Minimum TCE $\eps > 0$\\
  \hline
  Decision-aware sigmoid scaling \cite{vodrahalli2022uncalibrated} & Binary classification & Scalar confidence & Difference in final accuracy, human confidence in correct decision, rate of human decision change in response to AI advice & Train human behavior model in response to AI advice (human behavior training data required), calibrate AI output to minimize cross-entropy loss of human classification decision & $\alpha, \beta \ge 0$ constants for scaling and shifting AI confidence\\
  \hline
  Decision-aware conformal prediction \cite{straitouri2023improving} & Multi-class classification & Conformal set & Decision-maker probability of error & Given known or estimated confusion matrix of decision-maker $\delta$, optimization algorithm for optimal $\alpha$ to minimize decision error probability (requires calibration set) & Conformal error probability $0 < \alpha < 1$\\
  \hline
  P+L (probability and human labeler) \cite{kerrigan2021combining} & Multi-class classification & Label distribution & Classification error rate, ECE, classwise ECE, NLL & Post-hoc AI model calibration (requires calibration set), estimate human confusion matrix (requires human labels), combine with Bayes rule & Dirichlet prior parameters $\gamma > 0$, $\beta > 0$; Gaussin prior $\cN(\mu, \sigma^2)$ on $\log T$ for temperature scaling\\
  \hline
  Optimizing joint decision utility \cite{bansal2021is} & Multi-class classification & Class label distribution & Accuracy, expected utility, empirical utility & Minimize expected utility over training set & Utility parameters $\beta, \lambda$; human accuracy $a \in [0,1]$\\
  \hline
\end{tabular}}}

%% file: sections/notional_examples.tex
\section{Notional examples}
\label{sec:notional_examples}

In this section, we briefly illustrate how our self-assessment selection and design methodology might be applied to two realistic scenarios of national interest where AI transparency is crucial in informing downstream decision-making. To demonstrate the breadth of our approach to many problem settings, we discuss examples with human and autonomous decision-makers (i.e., $\delta$), which observe self-assessment outputs to make downstream decisions. We first describe a disaster relief scenario with a human decision-maker, followed by a UAV ISR tracking application with an autonomous decision-maker.

\subsection{Disaster relief triaging (human decision-maker)}
\label{subsec:triage}
\begin{figure}[htb]
\centering
\hspace*{\fill}%
\begin{subfigure}[t]{.5\textwidth}
    \centering
    \includegraphics[height=1.7in]{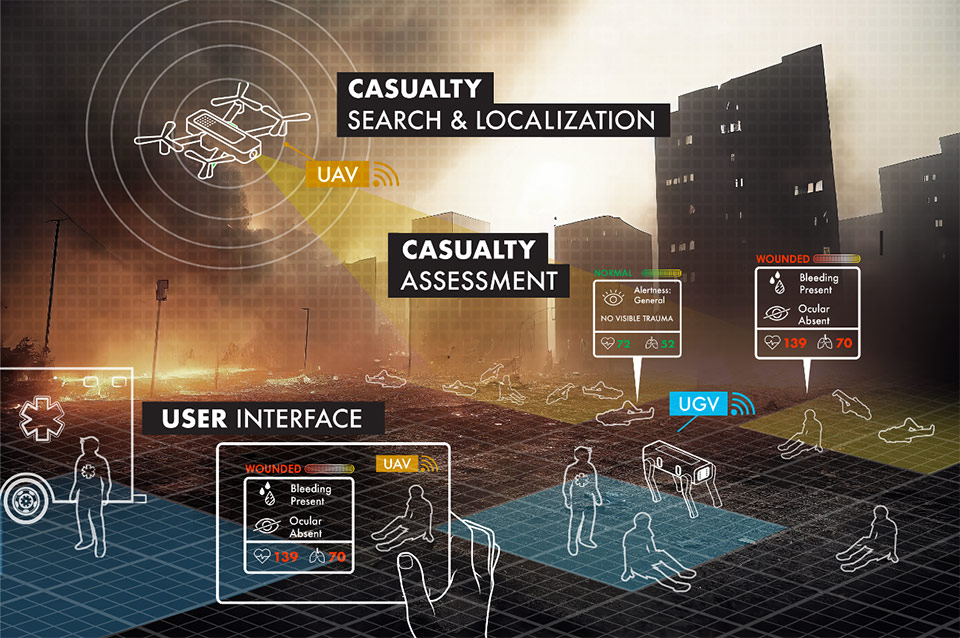}
    \caption{}
    \label{fig:triage:a}
\end{subfigure}%
\hfill%
\begin{subfigure}[t]{.5\textwidth}
    \centering
    \includegraphics[height=1.7in]{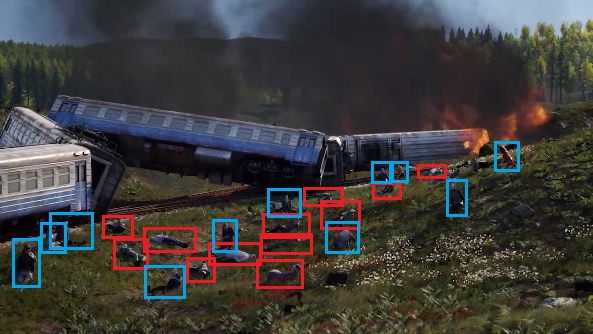}
    \caption{}
    \label{fig:triage:b}
\end{subfigure}%
\hspace*{\fill}%
\caption{(a) In settings similar to the DARPA Triage Challenge, autonomous agents are leveraged to assess medical triaging at scale. (b) In our notational use case, we envision an object-detection based triaging system to classify casualties based on predicted severity level. Images adapted and modified from \url{https://triagechallenge.darpa.mil/index}.}
\label{fig:triage}
\end{figure}

Medical \emph{triaging} during disaster relief scenarios is key to effectively applying limited medical resources to prioritize various casualty needs appropriately. As an example of a realistic disaster relief scenario and triaging effort, we consider a scenario inspired by the DARPA Triage Challenge\footnote{Further information on the challenge is available at \url{https://triagechallenge.darpa.mil/index}.} (DTC), which breaks triaging into two stages of primary and secondary triaging: 
\begin{itemize}
    \item \emph{Primary triaging} consists of identifying the severity of a situation, the presence of casualties, injury severity levels, and who requires hands-on medical evaluation or interventions.
    \item \emph{Secondary triaging} involves placing non-invasive sensors on these casualties to determine critical vs.\ non-critical conditions.
\end{itemize}
DARPA provides various datasets to tackle primary and secondary triaging. These range from Uncrewed Aircraft Vehicle (UAV) and Uncrewed Ground Vehicle (UGV) imagery of high-fidelity simulated scenarios, to purely tabular casualty medical information (\Cref{fig:triage:a}). At first glance, this complex, data-rich scenario presents a prime opportunity for AI algorithm development and automatic resource optimization. However, applying medical resources to casualties is inherently safety-critical, and therefore any use of AI in this setting must be accompanied by appropriate SeA, to increase overall trust in medical decisions being made. Furthermore, to maintain a human-in-the-loop in this safety-critical setting, we assume that a human operator makes the final decisions after receiving the AI's predictions and SeA, rather than having medical decisions being fully automated.

\subsubsection{Problem setup}
\label{subsec:Triage_setup}

In the notional disaster triaging scenario we consider, the primary mission is to maximize the number of saved lives by appropriately deploying resources to casualties present in the scene. We envision this task being performed by a human deciding where to deploy medical resources and personnel based on the outputs of an Object Detection and Classification model (stylization in \Cref{fig:triage:b}). Specifically, we assume the model receives as input high-resolution UAV RGB imagery captured by an expert operator (each image instance denoted by $x$ in our framework in \Cref{fig:notation}). We assume that the images will cover the entire incident area, ensuring all viewable casualties are within at least one frame and no frames are overlapping.

Each casualty is associated with a ``ground-truth'' class label ($y$) given by one of four states, corresponding to various levels of injury severity: \texttt{Healthy},\footnote{Here we slightly abuse the term ``casualty'' by allowing for the possibility that an identified person within the scene is actually unharmed.} \texttt{Delayed}, \texttt{Immediate}, or \texttt{Expectant}. These labels are a simplified mapping from recorded ground-truth labels, which include Trauma, Alertness, and Vitals presence/continuous values alongside the presence of critical factors, Severe Hemorrhaging and Respiratory distress. For this notional example, we assume there is a meaningful mapping from these health descriptors to injury severity. The Object Detection algorithm will determine each casualty location, while the Classifier predicts casualty state ($\widehat{y}$). We assume the forward model ($f_{\theta}$) is a combined Object Detection and Classifier model (e.g., YOLO \cite{redmon2016you}) that outputs as $z$ bounding box coordinates and a probability vector $p(x)$ containing the confidence of each class.\footnote{For the sake of illustration, we assume that the Object Detector's localization performance is flawless and that classification is the only task requiring confidence assessments.} As feedback to the human decision-maker, we assume these outputs are visualized by superimposing the output bounding boxes with their corresponding predicted injury state. We also assume that the visualization allows for a single scalar confidence value to be presented with each prediction.

Based on this visualization, the human decision-maker ($\delta$) then takes one of three possible actions ($a$), per casualty bounding box: \texttt{no action}, \texttt{deploy medical personnel}, \texttt{evacuate}. Each of these actions requires a different level of resource utilization, which can be interpreted as a literal ``cost'': the lack of an action does not deplete any resources while deploying medical personnel or ordering an evacuation requires increasing allocated resources. Furthermore, each action is potentially associated with an ``opportunity cost,'' depending on the ground-truth injury state of each casualty. For instance, deploying medical personnel to an identified casualty who is actually healthy may constitute a waste of resources, while failing to evacuate someone with a severe injury may result in a critical failure to save that person's life. We combine these various notions of ``cost'' into a single cost function $\ell(y, a)$, delineating the decision cost of all class-decision pairs (\Cref{table:triage_matrix}). Designing an appropriate cost matrix is a non-trivial challenge that should be undertaken by a domain expert; here, for the sake of illustration, we assign notional values to each state-action pair to capture the various degrees of severity in making ``incorrect'' decisions. Each casualty bounding box is associated with its own cost value $\ell(y,a)$ depending on the true state of the person in question and the ultimate action the downstream decision-maker takes.

\begin{table}[htb]
\centering
\input{figures/triage_matrix}
\caption{Cost matrix for disaster triaging scenario, depicting the decision cost $\ell(y,a)$ over all state-action pairs. In this context, absolute magnitudes do not carry meaning, but relative magnitudes represent the severity of ``incorrect'' decisions, where lower values are more desirable and higher values correspond to more severe consequences.}
\label{table:triage_matrix}
\end{table}




\subsubsection{Utilizing self-assessment guidance}

We now illustrate a notional process, utilizing the diagram in \Cref{fig:flowchart}, of identifying an appropriate self-assessment technique to aid a human decision-maker in taking actions that minimize the overall cost incurred when summed across casualties. Starting at the top of the flowchart, we identify the AI task as multi-class classification with classification accuracy as the performance measure since the Classifier component of interest in the forward model assigns one of three injury states to each bounding box casualty. In this example, we assume the forward model is already trained, and therefore we should utilize a post-hoc self-assessment method (assuming that a domain-specific calibration dataset is available). Since this problem setting is safety-critical with well-defined notions of cost (at least, in our notional description), we would like to leverage a decision-aware SeA method. By design, we assume that the forward model already outputs an estimated class probability vector $p(x)$, and that the visualization to the decision-maker allows for each bounding box state prediction to display an associated confidence value. Therefore, we limit ourselves to methods that output a scalar confidence to conform to the visualization constraints and the decision-maker's expectations.

Based on these requirements, after studying the methods in \Cref{sec:overview} we identify \emph{temperature scaling} as an appropriate technique, where we denote $\widetilde{p} = \sigma(\frac{1}{T} \log p(x))$ as a temperature-scaled probability vector and let $g_T(p(x)) = \widetilde{p}_{\widehat{y}}$ be the temperature-scaled probability value associated with the model's prediction, constituting a single scalar confidence to be presented to the decision-maker. In this context, rather than seeking a temperature $T$ that minimizes a calibration metric such as ECE, we instead seek to minimize the expected incurred cost $\E[\ell(y,a)]$, where this expectation is taken over the input data $x$, true labels $y$, as well as the (possibly stochastic) actions $a$ taken by the decision maker. In general, the exact human-decision maker policy $\delta$ given an image, predicted injury state, and SeA confidence will be unknown, preventing the direct optimization of this expected cost over $T$. However, for the sake of this example we will assume that an approximate model $\widehat{\delta}(\widehat{y}, g)$ of human behavior is available (as a function of prediction $\widehat{y}$ and self-assessment confidence $g$), which could possibly be fit to a dataset of recorded human decisions. With this approximate decision-maker policy available, we propose to directly optimize $\min_T \: E_{x,y \sim \cD_{X,Y}} \, \E_{a \sim \widehat{\delta}(\widehat{y}, g_T(p(x)))} [\ell(y, a)]$ to arrive at a selected temperature value $T$ for temperature scaling. While only notional, this process illustrates the utility of a formal approach for SeA selection in the context of a high-impact decision-driven problem.

\subsection{Autonomous UAV ISR (algorithmic decision maker)}
UAV-based Intelligence, Surveillance, and Reconnaissance (ISR) systems may sometimes benefit from an algorithmic decision-maker to autonomously complete a mission without explicit interventions or control from a human operator. However, decisions made by such algorithms can have varied consequences, especially in high-impact scenarios such as the battlespace. The potential consequences for certain actions, such as scanning an area of interest, may be mild (e.g., the worst-case outcome being increased fuel usage on an inefficient flight path that doesn't produce reportable intel), while other actions --- such as tracking a specific entity --- may pose greater levels of risk. Therefore, it is important for any AI tools utilized by an algorithmic decision-maker to be equipped with self-assessment, such that the decision algorithm can properly weigh the likelihood of various outcomes against their associated costs. Since different actions may have various levels of risk, it is important for SeA design in this context to explicitly take downstream algorithmic decisions into consideration. 

\subsubsection{Problem setup}
\label{subsec:uav_setup}

\begin{figure}[htb]
\centering
\includegraphics[width=0.6\textwidth]{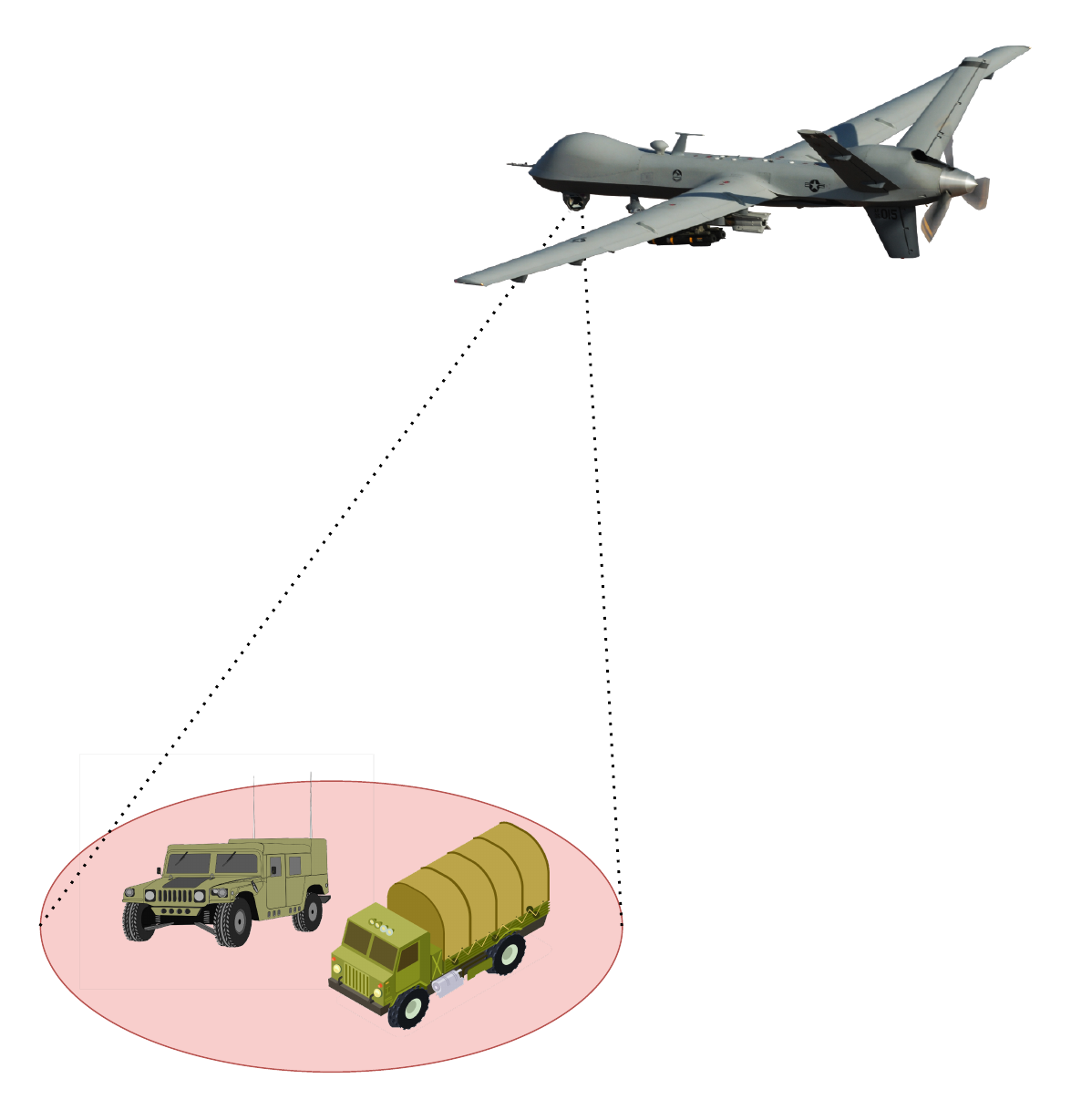}
\caption{Stylization of UAV vehicle detection and following.}
\label{fig:uav}
\end{figure}

In our notional example, we consider a scenario where an Autonomous UAV's mission is to find and follow adversary military vehicles (see stylization\footnote{UAV source material adapted from public domain imagery at \url{https://www.af.mil/News/Photos/igphoto/2000608254/mediaid/5461083/} (U.S.\ Air Force Photo / Lt.\ Col.\ Leslie Pratt). The appearance of U.S.\ Department of Defense (DoD) visual information does not imply or constitute DoD endorsement.} in \Cref{fig:uav}). We envision a scenario where the UAV uses onboard Electro-Optical (EO) sensors to detect and classify potential vehicles to follow. In this simplified example, we assume that each vehicle belongs to one of three classes ($y$): \texttt{friendly military vehicle}, \texttt{adversary military vehicle}, or 
\texttt{civilian vehicle}. For simplicity, we assume there will be exactly one vehicle of interest  driving through the UAV's field of view at a time. As in \Cref{subsec:triage}, we assume that an onboard AI system receives as input a single frame of RGB imagery ($x$, in our framework) and that the forward model ($f_{\theta}$) is a trained Object Detector that outputs ($z$) bounding boxes around the vehicle and includes a Classifier that outputs a class prediction $\widehat{y}$ along with a probability vector $g(x)$ estimating the likelihood of each class. For simplicity we assume that the Object Detector's localization performance is flawless, such that classification is the only task requiring self-assessment.

We assume that the onboard system toggles between two action modes: \emph{scanning} and \emph{following}. ``Scanning'' patrols an area in search of vehicles of possible interest while ``following'' pursues a target within the mission parameters. The downstream autonomous decision-maker will decide between these two actions based on the predicted class for each vehicle of interest. As in the disaster triage example, we describe a notional cost matrix in \Cref{table:uav_matrix} depicting a cost value $\ell(y, a)$ for each vehicle class-action pair that reflects the risks associated with each possibility (note that as in \Cref{subsec:triage}, only the relative differences between costs matter in this context, rather than the absolute scale). For example, we assign a cost of 50 to following a friendly military vehicle since this may constitute a waste of resources, but following a civilian is given a higher cost of 75 as this could be considered problematic. Deciding to ``scan area'' when an adversary military vehicle is actually in frame receives the highest loss of 100, which would be considered a mission failure. All other actions are considered ``correct'' and given a loss of 0. To choose scanning or following, an algorithmic decision-maker onboard the UAV (denoted by decision policy $\delta$) expects as input a probability vector $g$ over the three possible classes, and takes a corresponding action. Specifically, we assume that $\delta$ is given by the Bayes' optimal policy, i.e., select the action $a$ that minimizes the expected decision cost for the probabilities in $g$ as $a = \argmin_{a \in \{\mathrm{scan}, \mathrm{follow}\}} \frac{1}{3} \sum_{y'} g(y') \ell(y', a)$. Hence, the probability vector $g$ provided to this automated decision maker has a tremendous effect on the subsequent action taken.

\begin{table}[htb]
\centering
\input{figures/uav_matrix}
\caption{Cost matrix for the Autonomous UAV scenario. Absolute magnitudes do not carry meaning, but relative magnitudes represent the importance of ``incorrect'' decisions.}
\label{table:uav_matrix}
\end{table}



\subsubsection{Utilizing self-assessment guidance}
\label{subsec:UAV_util}

As in \Cref{subsec:triage}, we can follow the guidance in \Cref{fig:flowchart} to determine an appropriate self-assessment technique. In this notional example, the flowchart process is identical to that of the disaster triaging example until the question of SeA conforming to decision-maker expectations. Instead of expecting a scalar confidence, in this ISR example we assume that the algorithmic decision-maker expects to receive a full probability vector $g$ over the three possible classes for each vehicle. Examining the decision-aware SeA methods in \Cref{table:decision_aware} (since we would like to use a SeA method aware of downstream decision costs), if we assume that a calibration set is available\footnote{This is a reasonable assumption since before deploying a UAV-based ISR system it is conceivable to benchmark it on calibration runs.} we can utilize the Decision Calibration technique in \cite{zhao2021calibrating}, which is specifically designed for Bayes' optimal decision-makers. This method utilizes an iterative re-calibration algorithm to ensure that the expected cost (as computed using $g$) is close to the true cost (in expectation over the true data distribution).

%% file: figures/triage_matrix.tex
\begin{tabular}{ |c c||c|c|c| } 
 \hline
 \multicolumn{2}{|c||}{Cost Matrix}&\multicolumn{3}{c|}{Action ($a$)} \\
 \multicolumn{2}{|c||}{} & no action & deploy medical personnel & evacuate \\ 
 \hline
 \hline

 \multirow{4}{3em}{True State ($y$)} 
 & healthy     & 0   & 30 & 100 \\
 & delayed     & 50  & 0  & 30 \\
 & immediate   & 100 & 5  & 0 \\
 & expectant   & 50  & 10 & 20 \\
 \hline
\end{tabular}

%% file: figures/uav_matrix.tex
\begin{tabular}{ |c c||c|c| } 
 \hline
 \multicolumn{2}{|c||}{Cost Matrix}&\multicolumn{2}{c|}{Action $a$} \\
 \multicolumn{2}{|c||}{} & Follow & Scan Area \\ 
 \hline
 \hline

 \multirow{4}{4em}{True Class $y$} 
 & friendly military  & 50  & 0   \\
 & adversary military & 0   & 100 \\
 & civilian           & 75  & 0   \\
 \hline
\end{tabular}

%% file: sections/discussion.tex
\section{Discussion}
\label{sec:discussion}

In this manuscript, we have presented a methodology for designing and selecting self-assessment techniques to increase the trustworthiness and transparency of AI models, emphasizing methods driven by uncertainty quantification and awareness of how self-assessment outputs will ultimately be used in downstream decision-making. This manuscript has established an initial framework and formulation to select between and compare such self-assessment methods. While self-assessment techniques continue to be developed rapidly, the framework we establish here is general enough to categorize novel techniques according to the process outlined in \Cref{fig:flowchart}.

The suite of methods presented in \Cref{sec:overview} serves as a representative set of decision-agnostic and decision-aware approaches, covering a range of uncertainty representations, generic metrics, and estimation mechanisms. Many of these methods are complementary and emphasize various components of the AI-driven decision-making pipeline, from the AI outputs to models for the downstream decision-maker and incorporation of anticipated decision costs. Even so, many self-assessment techniques discussed here can still make unrealistic assumptions about the information available for self-assessment selection and optimization, such as having complete knowledge of the downstream decision cost function $\ell$ or decision-maker behavior $\delta$. Below we describe several core challenges to help bridge the gap between theoretical decision-driven self-assessment techniques and the next generation of practical self-assessment systems.

\paragraph{Decision-maker estimation} While it may be the case that the decision-maker's action policy $\delta$ is known exactly or approximately in certain scenarios --- such as with well-characterized human decision-makers or autonomous decision-makers with mathematically well-understood behaviors such as Kalman-filter based tracking algorithms --- in many scenarios, the decision function $\delta$ is unlikely to be known a priori. Furthermore, it is highly unlikely that realistic decision makers act rationally and Bayes' optimally, as is assumed in several self-assessment techniques (e.g., \cite{zhao2021calibrating}). Therefore, there is a crucial need for decision-aware self-assessment methods that do not rely on accurate models of the decision-making policy $\delta$, or instead learn such models ``on the fly'' at deployment. One promising avenue in this direction is described in \textcite{straitouri2023designing}, which does not assume knowledge of $\delta$ and instead takes an online learning approach to optimize self-assessment parameters based on ``in the loop'' decision-maker feedback.

\paragraph{Unknown decision costs} To our knowledge, the current set of available decision-aware self-assessment techniques all require knowledge of the downstream decision cost function $\ell(x,y,a)$ assessing the cost incurred by each decision-maker action (informed by AI self-assessment). In practice, it may not always be the case that such a cost function is known explicitly, and may only be accessible implicitly by observing the incurred costs made by each decision at test time, in which case online learning will likely be advantageous as a self-assessment tuning paradigm. Alternatively, it may be that \emph{implicit} knowledge of $\ell$ is available in preference judgments between actions made by an auxiliary oracle. In this case, principles from learning from human preferences and reward learning could potentially be deployed to estimate a surrogate cost $\widehat{\ell}$ from such relative feedback (see, for example, \cite{myer2023active}).

\paragraph{Limited calibration data} Many self-assessment methods (both decision-agnostic and decision-aware) require access to a held-out calibration dataset for fitting self-assessment parameters and choosing algorithm hyperparameters. However, in many practical scenarios there may not be the luxury of having such a large validation dataset available. This would be problematic in self-assessment techniques with heavy data requirements, such as \textcite{vodrahalli2022uncalibrated}, which first constructs an explicit model for a decision-maker $\delta$, based on behavioral response data. To mitigate such data costs, data-efficient techniques such as \emph{active learning} \cite{settles2009active} could potentially be deployed to only solicit and learn from the most informative test-time points. This judicious data selection could be deployed at multiple training stages, including learning the forward model $f$, self-assessment $g$, decision-maker $\delta$, and loss function $\ell$.

\paragraph{Multi-use self-assessment} In this manuscript, we have described self-assessment in the context of a single decision-maker $\delta$ and loss function $\ell$. However, in real-world scenarios these quantities may be inaccessible during self-assessment selection and tuning or may only be available during deployment. Nevertheless, it may be the case that $\delta$ and $\ell$ are known to belong to \emph{classes} of decision-makers and downstream cost functions, respectively. In this case, self-assessment should be selected and tuned to be compatible with any decision-maker or downstream cost function within these options. Initial strides were made to study self-assessment over classes of decision costs in \textcite{zhao2021calibrating}, but this formulation only applies to broad classes of losses under the assumption of Bayes' optimal decision makers.

\paragraph{}To our knowledge, there does not exist a single self-assessment method that addresses each of these practical challenges, which each needs to be addressed  in real-world self-assessment selection, tuning, and deployment. These methodology gaps present a prime opportunity for novel decision-driven self-assessment research, development, and evaluation.

%% file: main.bib
@InProceedings{redmon2016you,
author = {Redmon, Joseph and Divvala, Santosh and Girshick, Ross and Farhadi, Ali},
title = {You Only Look Once: Unified, Real-Time Object Detection},
booktitle = {Proceedings of the IEEE Conference on Computer Vision and Pattern Recognition (CVPR)},
month = {June},
year = {2016}
}

@InProceedings{cortes2016learning,
author="Cortes, Corinna
and DeSalvo, Giulia
and Mohri, Mehryar",
editor="Ortner, Ronald
and Simon, Hans Ulrich
and Zilles, Sandra",
title="Learning with Rejection",
booktitle="Algorithmic Learning Theory",
year="2016",
publisher="Springer International Publishing",
address="Cham",
pages="67--82",
isbn="978-3-319-46379-7"
}

@inproceedings{kumar2019verified,
 author = {Kumar, Ananya and Liang, Percy S and Ma, Tengyu},
 booktitle = {Advances in Neural Information Processing Systems},
 editor = {H. Wallach and H. Larochelle and A. Beygelzimer and F. d\textquotesingle Alch\'{e}-Buc and E. Fox and R. Garnett},
 pages = {},
 publisher = {Curran Associates, Inc.},
 title = {Verified Uncertainty Calibration},
 url = {https://proceedings.neurips.cc/paper_files/paper/2019/file/f8c0c968632845cd133308b1a494967f-Paper.pdf},
 volume = {32},
 year = {2019}
}

@INPROCEEDINGS{myer2023active,
  author={Myers, Vivek and Bıyık, Erdem and Sadigh, Dorsa},
  booktitle={2023 IEEE International Conference on Robotics and Automation (ICRA)}, 
  title={Active Reward Learning from Online Preferences}, 
  year={2023},
  volume={},
  number={},
  pages={7511-7518},
  doi={10.1109/ICRA48891.2023.10160439}}

@article{settles2009active,
  title={Active learning literature survey},
  author={Settles, Burr},
  year={2009},
  publisher={University of Wisconsin-Madison Department of Computer Sciences}
}

@article{dwivedi2023explainable,
author = {Dwivedi, Rudresh and Dave, Devam and Naik, Het and Singhal, Smiti and Omer, Rana and Patel, Pankesh and Qian, Bin and Wen, Zhenyu and Shah, Tejal and Morgan, Graham and Ranjan, Rajiv},
title = {Explainable AI (XAI): Core Ideas, Techniques, and Solutions},
year = {2023},
issue_date = {September 2023},
publisher = {Association for Computing Machinery},
address = {New York, NY, USA},
volume = {55},
number = {9},
issn = {0360-0300},
url = {https://doi.org/10.1145/3561048},
doi = {10.1145/3561048},
journal = {ACM Comput. Surv.},
month = {jan},
articleno = {194},
numpages = {33}
}

@Article{minh2022explainable,
author={Minh, Dang
and Wang, H. Xiang
and Li, Y. Fen
and Nguyen, Tan N.},
title={Explainable artificial intelligence: a comprehensive review},
journal={Artificial Intelligence Review},
year={2022},
month={Jun},
day={01},
volume={55},
number={5},
pages={3503-3568},
issn={1573-7462},
doi={10.1007/s10462-021-10088-y},
url={https://doi.org/10.1007/s10462-021-10088-y}
}

@inproceedings{
kirchenbauer2022what,
title={What is Your Metric Telling You? Evaluating Classifier Calibration under Context-Specific Definitions of Reliability},
author={John Kirchenbauer and Jacob R Oaks and Eric Heim},
booktitle={ML Evaluation Standards - ICLR 2022 Workshop},
year={2022},
url={https://ml-eval.github.io/assets/pdf/Calibration_Evaluation.pdf}
}

@inproceedings{nixon2019measuring,
  title={Measuring calibration in deep learning.},
  author={Nixon, Jeremy and Dusenberry, Michael W and Zhang, Linchuan and Jerfel, Ghassen and Tran, Dustin},
  booktitle={CVPR workshops},
  volume={2},
  number={7},
  year={2019}
}

@InProceedings{guo2017on,
  title = 	 {On Calibration of Modern Neural Networks},
  author =       {Chuan Guo and Geoff Pleiss and Yu Sun and Kilian Q. Weinberger},
  booktitle = 	 {Proceedings of the 34th International Conference on Machine Learning},
  pages = 	 {1321--1330},
  year = 	 {2017},
  editor = 	 {Precup, Doina and Teh, Yee Whye},
  volume = 	 {70},
  series = 	 {Proceedings of Machine Learning Research},
  month = 	 {06--11 Aug},
  publisher =    {PMLR},
  url = 	 {https://proceedings.mlr.press/v70/guo17a.html}
}

@inproceedings{
yona2022useful,
title={Useful Confidence Measures: Beyond the Max Score},
author={Gal Yona and Amir Feder and Itay Laish},
booktitle={NeurIPS 2022 Workshop on Distribution Shifts: Connecting Methods and Applications},
year={2022},
url={https://openreview.net/forum?id=CqUMx8sFhb}
}

@inproceedings{kull2019beyond,
 author = {Kull, Meelis and Perello Nieto, Miquel and K\"{a}ngsepp, Markus and Silva Filho, Telmo and Song, Hao and Flach, Peter},
 booktitle = {Advances in Neural Information Processing Systems},
 editor = {H. Wallach and H. Larochelle and A. Beygelzimer and F. d\textquotesingle Alch\'{e}-Buc and E. Fox and R. Garnett},
 pages = {},
 publisher = {Curran Associates, Inc.},
 title = {Beyond temperature scaling: Obtaining well-calibrated multi-class probabilities with Dirichlet calibration},
 url = {https://proceedings.neurips.cc/paper_files/paper/2019/file/8ca01ea920679a0fe3728441494041b9-Paper.pdf},
 volume = {32},
 year = {2019}
}

@inproceedings{rajendran2019accurate,
 author = {Rajendran, Vickram and LeVine, William},
 booktitle = {Advances in Neural Information Processing Systems},
 editor = {H. Wallach and H. Larochelle and A. Beygelzimer and F. d\textquotesingle Alch\'{e}-Buc and E. Fox and R. Garnett},
 pages = {},
 publisher = {Curran Associates, Inc.},
 title = {Accurate Layerwise Interpretable Competence Estimation},
 url = {https://proceedings.neurips.cc/paper_files/paper/2019/file/a11da6bd58b95b334f8cd49f00918f16-Paper.pdf},
 volume = {32},
 year = {2019}
}

@article{yang2024generalized,
  title={Generalized out-of-distribution detection: A survey},
  author={Yang, Jingkang and Zhou, Kaiyang and Li, Yixuan and Liu, Ziwei},
  journal={International Journal of Computer Vision},
  pages={1--28},
  year={2024},
  publisher={Springer}
}

@article{abdar2021review,
title = {A review of uncertainty quantification in deep learning: Techniques, applications and challenges},
journal = {Information Fusion},
volume = {76},
pages = {243-297},
year = {2021},
issn = {1566-2535},
doi = {https://doi.org/10.1016/j.inffus.2021.05.008},
url = {https://www.sciencedirect.com/science/article/pii/S1566253521001081},
author = {Moloud Abdar and Farhad Pourpanah and Sadiq Hussain and Dana Rezazadegan and Li Liu and Mohammad Ghavamzadeh and Paul Fieguth and Xiaochun Cao and Abbas Khosravi and U. Rajendra Acharya and Vladimir Makarenkov and Saeid Nahavandi}
}

@Article{gawlikowski2023survey,
author={Gawlikowski, Jakob
and Tassi, Cedrique Rovile Njieutcheu
and Ali, Mohsin
and Lee, Jongseok
and Humt, Matthias
and Feng, Jianxiang
and Kruspe, Anna
and Triebel, Rudolph
and Jung, Peter
and Roscher, Ribana
and Shahzad, Muhammad
and Yang, Wen
and Bamler, Richard
and Zhu, Xiao Xiang},
title={A survey of uncertainty in deep neural networks},
journal={Artificial Intelligence Review},
year={2023},
month={Oct},
day={01},
volume={56},
number={1},
pages={1513-1589},
issn={1573-7462},
doi={10.1007/s10462-023-10562-9},
url={https://doi.org/10.1007/s10462-023-10562-9}
}

@misc{ghosh2021uncertainty,
      title={Uncertainty Quantification 360: A Holistic Toolkit for Quantifying 
      and Communicating the Uncertainty of AI}, 
      author={Soumya Ghosh and Q. Vera Liao and Karthikeyan Natesan Ramamurthy 
      and Jiri Navratil and Prasanna Sattigeri 
      and Kush R. Varshney and Yunfeng Zhang},
      year={2021},
      eprint={2106.01410},
      archivePrefix={arXiv},
      primaryClass={cs.AI}
}

@article{chung2021uncertainty,
  title={Uncertainty Toolbox: an Open-Source Library for Assessing, Visualizing, and Improving Uncertainty Quantification},
  author={Chung, Youngseog and Char, Ian and Guo, Han and Schneider, Jeff and Neiswanger, Willie},
  journal={arXiv preprint arXiv:2109.10254},
  year={2021}
}

@article{naeini2015obtaining, title={Obtaining Well Calibrated Probabilities Using Bayesian Binning}, volume={29}, url={https://ojs.aaai.org/index.php/AAAI/article/view/9602}, DOI={10.1609/aaai.v29i1.9602}, abstractNote={ &lt;p&gt; Learning probabilistic predictive models that are well calibrated is critical for many prediction and decision-making tasks in artificial intelligence. In this paper we present a new non-parametric calibration method called Bayesian Binning into Quantiles (BBQ) which addresses key limitations of existing calibration methods. The method post processes the output of a binary classification algorithm; thus, it can be readily combined with many existing classification algorithms. The method is computationally tractable, and empirically accurate, as evidenced by the set of experiments reported here on both real and simulated datasets. &lt;/p&gt; }, number={1}, journal={Proceedings of the AAAI Conference on Artificial Intelligence}, author={Pakdaman Naeini, Mahdi and Cooper, Gregory and Hauskrecht, Milos}, year={2015}, month={Feb.} }

@InProceedings{gupta2021distribution,
  title = 	 {Distribution-Free Calibration Guarantees for Histogram Binning without Sample Splitting},
  author =       {Gupta, Chirag and Ramdas, Aaditya},
  booktitle = 	 {Proceedings of the 38th International Conference on Machine Learning},
  pages = 	 {3942--3952},
  year = 	 {2021},
  editor = 	 {Meila, Marina and Zhang, Tong},
  volume = 	 {139},
  series = 	 {Proceedings of Machine Learning Research},
  month = 	 {18--24 Jul},
  publisher =    {PMLR},
  url = 	 {https://proceedings.mlr.press/v139/gupta21b.html}
}

@inproceedings{zadrozny2001obtaining,
author = {Zadrozny, Bianca and Elkan, Charles},
title = {Obtaining calibrated probability estimates from decision trees and naive Bayesian classifiers},
year = {2001},
isbn = {1558607781},
publisher = {Morgan Kaufmann Publishers Inc.},
address = {San Francisco, CA, USA},
booktitle = {Proceedings of the Eighteenth International Conference on Machine Learning},
pages = {609–616},
numpages = {8},
series = {ICML '01}
}

@InProceedings{dheur2023large,
  title = 	 {A Large-Scale Study of Probabilistic Calibration in Neural Network Regression},
  author =       {Dheur, Victor and Ben Taieb, Souhaib},
  booktitle = 	 {Proceedings of the 40th International Conference on Machine Learning},
  pages = 	 {7813--7836},
  year = 	 {2023},
  editor = 	 {Krause, Andreas and Brunskill, Emma and Cho, Kyunghyun and Engelhardt, Barbara and Sabato, Sivan and Scarlett, Jonathan},
  volume = 	 {202},
  series = 	 {Proceedings of Machine Learning Research},
  month = 	 {23--29 Jul},
  publisher =    {PMLR},
  url = 	 {https://proceedings.mlr.press/v202/dheur23a.html}
}

@InProceedings{dheur2024probabilistic,
  title = 	 { Probabilistic Calibration by Design for Neural Network Regression },
  author =       {Dheur, Victor and Ben Taieb, Souhaib},
  booktitle = 	 {Proceedings of The 27th International Conference on Artificial Intelligence and Statistics},
  pages = 	 {3133--3141},
  year = 	 {2024},
  editor = 	 {Dasgupta, Sanjoy and Mandt, Stephan and Li, Yingzhen},
  volume = 	 {238},
  series = 	 {Proceedings of Machine Learning Research},
  month = 	 {02--04 May},
  publisher =    {PMLR},
  url = 	 {https://proceedings.mlr.press/v238/dheur24a.html}
}

@InProceedings{bui2024density,
  title = 	 { Density-Regression: Efficient and Distance-aware Deep Regressor for Uncertainty Estimation under Distribution Shifts },
  author =       {Manh Bui, Ha and Liu, Anqi},
  booktitle = 	 {Proceedings of The 27th International Conference on Artificial Intelligence and Statistics},
  pages = 	 {2998--3006},
  year = 	 {2024},
  editor = 	 {Dasgupta, Sanjoy and Mandt, Stephan and Li, Yingzhen},
  volume = 	 {238},
  series = 	 {Proceedings of Machine Learning Research},
  month = 	 {02--04 May},
  publisher =    {PMLR},
  url = 	 {https://proceedings.mlr.press/v238/manh-bui24a.html}
}

@inproceedings{capone2023sharp,
 author = {Capone, Alexandre and Hirche, Sandra and Pleiss, Geoff},
 booktitle = {Advances in Neural Information Processing Systems},
 editor = {A. Oh and T. Naumann and A. Globerson and K. Saenko and M. Hardt and S. Levine},
 pages = {36579--36590},
 publisher = {Curran Associates, Inc.},
 title = {Sharp Calibrated Gaussian Processes},
 url = {https://proceedings.neurips.cc/paper_files/paper/2023/file/7319b7561ffe5e2f6419acd4a2f52d6b-Paper-Conference.pdf},
 volume = {36},
 year = {2023}
}

@InProceedings{popordanoska2024beyond,
    author    = {Popordanoska, Teodora and Tiulpin, Aleksei and Blaschko, Matthew B.},
    title     = {Beyond Classification: Definition and Density-Based Estimation of Calibration in Object Detection},
    booktitle = {Proceedings of the IEEE/CVF Winter Conference on Applications of Computer Vision (WACV)},
    month     = {January},
    year      = {2024},
    pages     = {585-594}
}

@InProceedings{oksuz2023towards,
    author    = {Oksuz, Kemal and Joy, Tom and Dokania, Puneet K.},
    title     = {Towards Building Self-Aware Object Detectors via Reliable Uncertainty Quantification and Calibration},
    booktitle = {Proceedings of the IEEE/CVF Conference on Computer Vision and Pattern Recognition (CVPR)},
    month     = {June},
    year      = {2023},
    pages     = {9263-9274}
}

@InProceedings{munir2023bridging,
    author    = {Munir, Muhammad Akhtar and Khan, Muhammad Haris and Khan, Salman and Khan, Fahad Shahbaz},
    title     = {Bridging Precision and Confidence: A Train-Time Loss for Calibrating Object Detection},
    booktitle = {Proceedings of the IEEE/CVF Conference on Computer Vision and Pattern Recognition (CVPR)},
    month     = {June},
    year      = {2023},
    pages     = {11474-11483}
}

@InProceedings{pathiraja2023multiclass,
    author    = {Pathiraja, Bimsara and Gunawardhana, Malitha and Khan, Muhammad Haris},
    title     = {Multiclass Confidence and Localization Calibration for Object Detection},
    booktitle = {Proceedings of the IEEE/CVF Conference on Computer Vision and Pattern Recognition (CVPR)},
    month     = {June},
    year      = {2023},
    pages     = {19734-19743}
}

@inproceedings{lakshminarayanan2017simple,
 author = {Lakshminarayanan, Balaji and Pritzel, Alexander and Blundell, Charles},
 booktitle = {Advances in Neural Information Processing Systems},
 editor = {I. Guyon and U. Von Luxburg and S. Bengio and H. Wallach and R. Fergus and S. Vishwanathan and R. Garnett},
 pages = {},
 publisher = {Curran Associates, Inc.},
 title = {Simple and Scalable Predictive Uncertainty Estimation using Deep Ensembles},
 url = {https://proceedings.neurips.cc/paper_files/paper/2017/file/9ef2ed4b7fd2c810847ffa5fa85bce38-Paper.pdf},
 volume = {30},
 year = {2017}
}

@InProceedings{gal2016dropout,
  title = 	 {Dropout as a Bayesian Approximation: Representing Model Uncertainty in Deep Learning},
  author = 	 {Gal, Yarin and Ghahramani, Zoubin},
  booktitle = 	 {Proceedings of The 33rd International Conference on Machine Learning},
  pages = 	 {1050--1059},
  year = 	 {2016},
  editor = 	 {Balcan, Maria Florina and Weinberger, Kilian Q.},
  volume = 	 {48},
  series = 	 {Proceedings of Machine Learning Research},
  address = 	 {New York, New York, USA},
  month = 	 {20--22 Jun},
  publisher =    {PMLR},
  url = 	 {https://proceedings.mlr.press/v48/gal16.html}
}

@inproceedings{osband2023epistemic,
 author = {Osband, Ian and Wen, Zheng and Asghari, Seyed Mohammad and Dwaracherla, Vikranth and IBRAHIMI, MORTEZA and Lu, Xiuyuan and Van Roy, Benjamin},
 booktitle = {Advances in Neural Information Processing Systems},
 editor = {A. Oh and T. Naumann and A. Globerson and K. Saenko and M. Hardt and S. Levine},
 pages = {2795--2823},
 publisher = {Curran Associates, Inc.},
 title = {Epistemic Neural Networks},
 url = {https://proceedings.neurips.cc/paper_files/paper/2023/file/07fbde96bee50f4e09303fd4f877c2f3-Paper-Conference.pdf},
 volume = {36},
 year = {2023}
}

@article{angelopoulos2023conformal,
url = {http://dx.doi.org/10.1561/2200000101},
year = {2023},
volume = {16},
journal = {Foundations and Trends® in Machine Learning},
title = {Conformal Prediction: A Gentle Introduction},
doi = {10.1561/2200000101},
issn = {1935-8237},
number = {4},
pages = {494-591},
author = {Anastasios N. Angelopoulos and Stephen Bates}
}

@InProceedings{kumar2018trainable,
  title = 	 {Trainable Calibration Measures for Neural Networks from Kernel Mean Embeddings},
  author =       {Kumar, Aviral and Sarawagi, Sunita and Jain, Ujjwal},
  booktitle = 	 {Proceedings of the 35th International Conference on Machine Learning},
  pages = 	 {2805--2814},
  year = 	 {2018},
  editor = 	 {Dy, Jennifer and Krause, Andreas},
  volume = 	 {80},
  series = 	 {Proceedings of Machine Learning Research},
  month = 	 {10--15 Jul},
  publisher =    {PMLR},
  url = 	 {https://proceedings.mlr.press/v80/kumar18a.html}
}

@inproceedings{kendall2017what,
 author = {Kendall, Alex and Gal, Yarin},
 booktitle = {Advances in Neural Information Processing Systems},
 editor = {I. Guyon and U. Von Luxburg and S. Bengio and H. Wallach and R. Fergus and S. Vishwanathan and R. Garnett},
 pages = {},
 publisher = {Curran Associates, Inc.},
 title = {What Uncertainties Do We Need in Bayesian Deep Learning for Computer Vision?},
 url = {https://proceedings.neurips.cc/paper_files/paper/2017/file/2650d6089a6d640c5e85b2b88265dc2b-Paper.pdf},
 volume = {30},
 year = {2017}
}

@inproceedings{maddox2019simple,
 author = {Maddox, Wesley J and Izmailov, Pavel and Garipov, Timur and Vetrov, Dmitry P and Wilson, Andrew Gordon},
 booktitle = {Advances in Neural Information Processing Systems},
 editor = {H. Wallach and H. Larochelle and A. Beygelzimer and F. d\textquotesingle Alch\'{e}-Buc and E. Fox and R. Garnett},
 pages = {},
 publisher = {Curran Associates, Inc.},
 title = {A Simple Baseline for Bayesian Uncertainty in Deep Learning},
 url = {https://proceedings.neurips.cc/paper_files/paper/2019/file/118921efba23fc329e6560b27861f0c2-Paper.pdf},
 volume = {32},
 year = {2019}
}

@InProceedings{wright2024analytic,
  title = 	 { An Analytic Solution to Covariance Propagation in Neural Networks },
  author =       {Wright, Oren and Nakahira, Yorie and M. F. Moura, Jos\'{e}},
  booktitle = 	 {Proceedings of The 27th International Conference on Artificial Intelligence and Statistics},
  pages = 	 {4087--4095},
  year = 	 {2024},
  editor = 	 {Dasgupta, Sanjoy and Mandt, Stephan and Li, Yingzhen},
  volume = 	 {238},
  series = 	 {Proceedings of Machine Learning Research},
  month = 	 {02--04 May},
  publisher =    {PMLR},
  url = 	 {https://proceedings.mlr.press/v238/wright24a.html}
}

@InProceedings{blundell2015weight,
  title = 	 {Weight Uncertainty in Neural Network},
  author = 	 {Blundell, Charles and Cornebise, Julien and Kavukcuoglu, Koray and Wierstra, Daan},
  booktitle = 	 {Proceedings of the 32nd International Conference on Machine Learning},
  pages = 	 {1613--1622},
  year = 	 {2015},
  editor = 	 {Bach, Francis and Blei, David},
  volume = 	 {37},
  series = 	 {Proceedings of Machine Learning Research},
  address = 	 {Lille, France},
  month = 	 {07--09 Jul},
  publisher =    {PMLR},
  url = 	 {https://proceedings.mlr.press/v37/blundell15.html}
}

@inproceedings{lipton2018detecting,
  title={Detecting and correcting for label shift with black box predictors},
  author={Lipton, Zachary and Wang, Yu-Xiang and Smola, Alexander},
  booktitle={International conference on machine learning},
  pages={3122--3130},
  year={2018},
  organization={PMLR}
}

@InProceedings{david2010impossibility,
  title = 	 {Impossibility Theorems for Domain Adaptation},
  author = 	 {David, Shai Ben and Lu, Tyler and Luu, Teresa and Pal, David},
  booktitle = 	 {Proceedings of the Thirteenth International Conference on Artificial Intelligence and Statistics},
  pages = 	 {129--136},
  year = 	 {2010},
  editor = 	 {Teh, Yee Whye and Titterington, Mike},
  volume = 	 {9},
  series = 	 {Proceedings of Machine Learning Research},
  address = 	 {Chia Laguna Resort, Sardinia, Italy},
  month = 	 {13--15 May},
  publisher =    {PMLR},
  url = 	 {https://proceedings.mlr.press/v9/david10a.html}
}

@inproceedings{
garg2022leveraging,
title={Leveraging unlabeled data to predict out-of-distribution performance},
author={Saurabh Garg and Sivaraman Balakrishnan and Zachary Chase Lipton and Behnam Neyshabur and Hanie Sedghi},
booktitle={International Conference on Learning Representations},
year={2022},
url={https://openreview.net/forum?id=o_HsiMPYh_x}
}

@ARTICLE{zhang2023survey,
  author={Zhang, Xu-Yao and Xie, Guo-Sen and Li, Xiuli and Mei, Tao and Liu, Cheng-Lin},
  journal={Proceedings of the IEEE}, 
  title={A Survey on Learning to Reject}, 
  year={2023},
  volume={111},
  number={2},
  pages={185-215},
  doi={10.1109/JPROC.2023.3238024}}

@inproceedings{liu2020energy,
 author = {Liu, Weitang and Wang, Xiaoyun and Owens, John and Li, Yixuan},
 booktitle = {Advances in Neural Information Processing Systems},
 editor = {H. Larochelle and M. Ranzato and R. Hadsell and M.F. Balcan and H. Lin},
 pages = {21464--21475},
 publisher = {Curran Associates, Inc.},
 title = {Energy-based Out-of-distribution Detection},
 url = {https://proceedings.neurips.cc/paper_files/paper/2020/file/f5496252609c43eb8a3d147ab9b9c006-Paper.pdf},
 volume = {33},
 year = {2020}
}

@inproceedings{
yang2024imagenetood,
title={ImageNet-{OOD}: Deciphering Modern Out-of-Distribution Detection Algorithms},
author={William Yang and Byron Zhang and Olga Russakovsky},
booktitle={The Twelfth International Conference on Learning Representations},
year={2024},
url={https://openreview.net/forum?id=VTYg5ykEGS}
}

@inproceedings{
hendrycks2017a,
title={A Baseline for Detecting Misclassified and Out-of-Distribution Examples in Neural Networks},
author={Dan Hendrycks and Kevin Gimpel},
booktitle={International Conference on Learning Representations},
year={2017},
url={https://openreview.net/forum?id=Hkg4TI9xl}
}

@inproceedings{yuksekgonul2023beyond,
 author = {Yuksekgonul, Mert and Zhang, Linjun and Zou, James Y and Guestrin, Carlos},
 booktitle = {Advances in Neural Information Processing Systems},
 editor = {A. Oh and T. Naumann and A. Globerson and K. Saenko and M. Hardt and S. Levine},
 pages = {38420--38453},
 publisher = {Curran Associates, Inc.},
 title = {Beyond Confidence: Reliable Models Should Also Consider Atypicality},
 url = {https://proceedings.neurips.cc/paper_files/paper/2023/file/7900318ffaf5e9bc60250f134c6cc3c7-Paper-Conference.pdf},
 volume = {36},
 year = {2023}
}

@inproceedings{marx2023calibration,
 author = {Marx, Charlie and Zalouk, Sofian and Ermon, Stefano},
 booktitle = {Advances in Neural Information Processing Systems},
 editor = {A. Oh and T. Naumann and A. Globerson and K. Saenko and M. Hardt and S. Levine},
 pages = {25910--25928},
 publisher = {Curran Associates, Inc.},
 title = {Calibration by Distribution Matching: Trainable Kernel Calibration Metrics},
 url = {https://proceedings.neurips.cc/paper_files/paper/2023/file/52493d82db00e73abb2858a5a5f28717-Paper-Conference.pdf},
 volume = {36},
 year = {2023}
}

@inproceedings{zhao2021calibrating,
 author = {Zhao, Shengjia and Kim, Michael and Sahoo, Roshni and Ma, Tengyu and Ermon, Stefano},
 booktitle = {Advances in Neural Information Processing Systems},
 editor = {M. Ranzato and A. Beygelzimer and Y. Dauphin and P.S. Liang and J. Wortman Vaughan},
 pages = {22313--22324},
 publisher = {Curran Associates, Inc.},
 title = {Calibrating Predictions to Decisions: A Novel Approach to Multi-Class Calibration},
 url = {https://proceedings.neurips.cc/paper_files/paper/2021/file/bbc92a647199b832ec90d7cf57074e9e-Paper.pdf},
 volume = {34},
 year = {2021}
}

@inproceedings{vodrahalli2022uncalibrated,
 author = {Vodrahalli, Kailas and Gerstenberg, Tobias and Zou, James Y},
 booktitle = {Advances in Neural Information Processing Systems},
 editor = {S. Koyejo and S. Mohamed and A. Agarwal and D. Belgrave and K. Cho and A. Oh},
 pages = {4004--4016},
 publisher = {Curran Associates, Inc.},
 title = {Uncalibrated Models Can Improve Human-AI Collaboration},
 url = {https://proceedings.neurips.cc/paper_files/paper/2022/file/1968ea7d985aa377e3a610b05fc79be0-Paper-Conference.pdf},
 volume = {35},
 year = {2022}
}

@InProceedings{straitouri2023improving,
  title = 	 {Improving Expert Predictions with Conformal Prediction},
  author =       {Straitouri, Eleni and Wang, Lequn and Okati, Nastaran and Gomez Rodriguez, Manuel},
  booktitle = 	 {Proceedings of the 40th International Conference on Machine Learning},
  pages = 	 {32633--32653},
  year = 	 {2023},
  editor = 	 {Krause, Andreas and Brunskill, Emma and Cho, Kyunghyun and Engelhardt, Barbara and Sabato, Sivan and Scarlett, Jonathan},
  volume = 	 {202},
  series = 	 {Proceedings of Machine Learning Research},
  month = 	 {23--29 Jul},
  publisher =    {PMLR},
  url = 	 {https://proceedings.mlr.press/v202/straitouri23a.html}
}

@inproceedings{sahoo2021reliable,
 author = {Sahoo, Roshni and Zhao, Shengjia and Chen, Alyssa and Ermon, Stefano},
 booktitle = {Advances in Neural Information Processing Systems},
 editor = {M. Ranzato and A. Beygelzimer and Y. Dauphin and P.S. Liang and J. Wortman Vaughan},
 pages = {1831--1844},
 publisher = {Curran Associates, Inc.},
 title = {Reliable Decisions with Threshold Calibration},
 url = {https://proceedings.neurips.cc/paper_files/paper/2021/file/0e65972dce68dad4d52d063967f0a705-Paper.pdf},
 volume = {34},
 year = {2021}
}

@article{bansal2021is, title={Is the Most Accurate AI the Best Teammate? Optimizing AI for Teamwork}, volume={35}, url={https://ojs.aaai.org/index.php/AAAI/article/view/17359}, DOI={10.1609/aaai.v35i13.17359}, abstractNote={AI practitioners typically strive to develop the most accurate systems, making an implicit assumption that the AI system will function autonomously. However, in practice, AI systems often are used to provide advice to people in domains ranging from criminal justice and finance to healthcare. In such AI-advised decision making, humans and machines form a team, where the human is responsible for making final decisions. But is the most accurate AI the best teammate? We argue &quot;not necessarily&quot; --- predictable performance may be worth a slight sacrifice in AI accuracy. Instead, we argue that AI systems should be trained in a human-centered manner, directly optimized for team performance. We study this proposal for a specific type of human-AI teaming, where the human overseer chooses to either accept the AI recommendation or solve the task themselves. To optimize the team performance for this setting we maximize the team’s expected utility, expressed in terms of the quality of the final decision, cost of verifying, and individual accuracies of people and machines. Our experiments with linear and non-linear models on real-world, high-stakes datasets show that the most accuracy AI may not lead to highest team performance and show the benefit of modeling teamwork during training through improvements in expected team utility across datasets, considering parameters such as human skill and the cost of mistakes. We discuss the shortcoming of current optimization approaches beyond well-studied loss functions such as log-loss, and encourage future work on AI optimization problems motivated by human-AI collaboration.}, number={13}, journal={Proceedings of the AAAI Conference on Artificial Intelligence}, author={Bansal, Gagan and Nushi, Besmira and Kamar, Ece and Horvitz, Eric and Weld, Daniel S.}, year={2021}, month={May}, pages={11405-11414} }

@inproceedings{kerrigan2021combining,
 author = {Kerrigan, Gavin and Smyth, Padhraic and Steyvers, Mark},
 booktitle = {Advances in Neural Information Processing Systems},
 editor = {M. Ranzato and A. Beygelzimer and Y. Dauphin and P.S. Liang and J. Wortman Vaughan},
 pages = {4421--4434},
 publisher = {Curran Associates, Inc.},
 title = {Combining Human Predictions with Model Probabilities via Confusion Matrices and Calibration},
 url = {https://proceedings.neurips.cc/paper_files/paper/2021/file/234b941e88b755b7a72a1c1dd5022f30-Paper.pdf},
 volume = {34},
 year = {2021}
}

@article{straitouri2023designing,
  title={Designing decision support systems using counterfactual prediction sets},
  author={Straitouri, Eleni and Rodriguez, Manuel Gomez},
  journal={arXiv preprint arXiv:2306.03928},
  year={2023}
}

@inproceedings{corvelo2023human,
 author = {Corvelo Benz, Nina and Rodriguez, Manuel},
 booktitle = {Advances in Neural Information Processing Systems},
 editor = {A. Oh and T. Naumann and A. Globerson and K. Saenko and M. Hardt and S. Levine},
 pages = {14609--14636},
 publisher = {Curran Associates, Inc.},
 title = {Human-Aligned Calibration for AI-Assisted Decision Making},
 url = {https://proceedings.neurips.cc/paper_files/paper/2023/file/2f1d1196426ba84f47d115cac3dcb9d8-Paper-Conference.pdf},
 volume = {36},
 year = {2023}
}

@article{cresswell2024conformal,
  title={Conformal Prediction Sets Improve Human Decision Making},
  author={Cresswell, Jesse C and Sui, Yi and Kumar, Bhargava and Vouitsis, No{\"e}l},
  journal={arXiv preprint arXiv:2401.13744},
  year={2024}
}

@inproceedings{babbar2022on,
  title     = {On the Utility of Prediction Sets in Human-AI Teams},
  author    = {Babbar, Varun and Bhatt, Umang and Weller, Adrian},
  booktitle = {Proceedings of the Thirty-First International Joint Conference on
               Artificial Intelligence, {IJCAI-22}},
  publisher = {International Joint Conferences on Artificial Intelligence Organization},
  editor    = {Lud De Raedt},
  pages     = {2457--2463},
  year      = {2022},
  month     = {7},
  note      = {Main Track},
  doi       = {10.24963/ijcai.2022/341},
  url       = {https://doi.org/10.24963/ijcai.2022/341},
}
